\documentclass[10pt,twocolumn,letterpaper]{article}

\PassOptionsToPackage{table}{xcolor}
\usepackage[pagenumbers]{iccv}       
\usepackage{tabularx}
\usepackage{multirow}
\usepackage[per-mode=symbol, group-digits=integer, detect-all=true, input-symbols={()}]{siunitx}
\usepackage{bm}       
\usepackage[percent]{overpic}
\usepackage{fp} 
\usepackage{graphicx}
\usepackage{tikz}
\usetikzlibrary{fit}
\usetikzlibrary{positioning}
\usetikzlibrary{shapes.geometric, shapes, arrows, arrows.meta, bending}
\usetikzlibrary{backgrounds}
\usetikzlibrary{spy}
\usepackage{pgfplots}
\pgfplotsset{compat=1.17}
\usepackage{fontawesome5}
\usepackage[export]{adjustbox}
\usepackage{bbding} 
\usepackage{bbm}
\usepackage[outline]{contour} 
\usepackage[table]{xcolor}
\contourlength{0.1em}

\newcommand*{\inparagraph}[1]{\smallskip\noindent\textbf{#1}\hspace{0.4em}}

\definecolor{iccvblue}{rgb}{0.21,0.49,0.74}
\definecolor{tud0d}{RGB}{83,83,83}
\definecolor{tud0c}{RGB}{137,137,137}
\definecolor{tud0b}{RGB}{181,181,181}
\definecolor{tud0a}{RGB}{220,220,220}
\definecolor{tud1a}{RGB}{93,133,195}
\definecolor{tud2a}{RGB}{0,156,218}
\definecolor{tud3a}{RGB}{80,182,149}
\definecolor{tud4a}{RGB}{175,204,80}
\definecolor{tud5a}{RGB}{221,223,72}
\definecolor{tud6a}{RGB}{255,224,92}
\definecolor{tud7a}{RGB}{248,186,60}
\definecolor{tud8a}{RGB}{238,122,52}
\definecolor{tud9a}{RGB}{233,80,62}
\definecolor{tud10a}{RGB}{201,48,142}
\definecolor{tud11a}{RGB}{128,69,151}
\definecolor{tud1b}{RGB}{0,90,169}
\definecolor{tud2b}{RGB}{0,131,204}
\definecolor{tud3b}{RGB}{0,157,129}
\definecolor{tud4b}{RGB}{153,192,0}
\definecolor{tud5b}{RGB}{201,212,0}
\definecolor{tud6b}{RGB}{253,202,0}
\definecolor{tud7b}{RGB}{245,163,0}
\definecolor{tud8b}{RGB}{236,101,0}
\definecolor{tud9b}{RGB}{230,0,26}
\definecolor{tud10b}{RGB}{166,0,132}
\definecolor{tud11b}{RGB}{114,16,133}
\definecolor{tud1c}{RGB}{0,78,138}
\definecolor{tud2c}{RGB}{0,104,157}
\definecolor{tud3c}{RGB}{0,136,119}
\definecolor{tud4c}{RGB}{127,171,22}
\definecolor{tud5c}{RGB}{177,189,0}
\definecolor{tud6c}{RGB}{215,172,0}
\definecolor{tud7c}{RGB}{210,135,0}
\definecolor{tud8c}{RGB}{204,76,3}
\definecolor{tud9c}{RGB}{185,15,34}
\definecolor{tud10c}{RGB}{149,17,105}
\definecolor{tud11c}{RGB}{97,28,115}
\definecolor{tud1d}{RGB}{36,53,114}
\definecolor{tud2d}{RGB}{0,78,115}
\definecolor{tud3d}{RGB}{0,113,94}
\definecolor{tud4d}{RGB}{106,139,55}
\definecolor{tud5d}{RGB}{153,166,4}
\definecolor{tud6d}{RGB}{174,142,0}
\definecolor{tud7d}{RGB}{190,111,0}
\definecolor{tud8d}{RGB}{169,73,19}
\definecolor{tud9d}{RGB}{156,28,38}
\definecolor{tud10d}{RGB}{115,32,84}
\definecolor{tud11d}{RGB}{76,34,106}
\definecolor{unlabeled}{RGB}{0,0,0}
\definecolor{egovehicle}{RGB}{0,0,0}
\definecolor{rectification border}{RGB}{0,0,0}
\definecolor{outofroi}{RGB}{0,0,0}
\definecolor{static}{RGB}{0,0,0}
\definecolor{dynamic}{RGB}{111,74,0}
\definecolor{ground}{RGB}{81,0,81}
\definecolor{road}{RGB}{128,64,128}
\definecolor{sidewalk}{RGB}{244,35,232}
\definecolor{parking}{RGB}{250,170,160}
\definecolor{rail track}{RGB}{230,150,140}
\definecolor{building}{RGB}{70,70,70}
\definecolor{wall}{RGB}{102,102,156}
\definecolor{fence}{RGB}{190,153,153}
\definecolor{guard rail}{RGB}{180,165,180}
\definecolor{bridge}{RGB}{150,100,100}
\definecolor{tunnel}{RGB}{150,120,90}
\definecolor{pole}{RGB}{153,153,153}
\definecolor{polegroup}{RGB}{153,153,153}
\definecolor{trafficlight}{RGB}{250,170,30}
\definecolor{trafficsign}{RGB}{220,220,0}
\definecolor{vegetation}{RGB}{107,142,35}
\definecolor{terrain}{RGB}{152,251,152}
\definecolor{sky}{RGB}{70,130,220}
\definecolor{skylight}{RGB}{98,182,252}
\definecolor{person}{RGB}{220,20,60}
\definecolor{rider}{RGB}{255,0,0}
\definecolor{car}{RGB}{0,0,142}
\definecolor{truck}{RGB}{0,0,70}
\definecolor{bus}{RGB}{0,60,100}
\definecolor{caravan}{RGB}{0,0,90}
\definecolor{trailer}{RGB}{0,0,110}
\definecolor{train}{RGB}{0,80,100}
\definecolor{motorcycle}{RGB}{0,0,230}
\definecolor{bicycle}{RGB}{119,11,32}
\definecolor{licenseplate}{RGB}{0,0,142}

\newcommand{\MethodName}{MR-DINOSAUR\@\xspace}

\usepackage{float}

\usepackage[pagebackref,breaklinks,colorlinks,allcolors=iccvblue]{hyperref}

\def\paperID{---} 
\def\confName{ICCVW}
\def\confYear{2025}

\title{Motion-Refined DINOSAUR for Unsupervised Multi-Object Discovery}

\newcommand{\authorstep}{\hspace{0.75cm}}
\newcommand{\affiliationstep}{\hspace{0.5cm}}

\author{
Xinrui Gong\textsuperscript{\normalfont{}* 1}
\authorstep Oliver Hahn\textsuperscript{\normalfont{}* 1}
\authorstep Christoph Reich\textsuperscript{\normalfont{}1,2,3,4}
\authorstep Krishnakant Singh\textsuperscript{\normalfont{} 1}\\
\authorstep Simone Schaub-Meyer\textsuperscript{\normalfont{} 1,5}
\authorstep Daniel Cremers\textsuperscript{\normalfont{} 2,3,4}
\authorstep Stefan Roth\textsuperscript{\normalfont{} 1,4,5}\\[1pt]
\small{\textsuperscript{1}TU Darmstadt \affiliationstep
\textsuperscript{2}TU Munich \affiliationstep \textsuperscript{3}MCML\affiliationstep \textsuperscript{4}ELIZA\affiliationstep \textsuperscript{5}hessian.AI\affiliationstep
\textsuperscript{*}equal contribution}\\ 
\small {\url{https://github.com/visinf/mr-dinosaur}}}

\usepackage{fancyhdr}
\usepackage{setspace}
\fancypagestyle{firststyle}
{

	\fancyhf{}
	\lfoot{{\footnotesize\begin{spacing}{.5}\parbox{\linewidth}{\vspace{-0.85em}%
					\textcolor{gray}{To appear in \emph{Proceedings of the IEEE/CVF International Conference on Computer Vision Workshops}, Honolulu, Hawai'i, USA, 2025.%
					\\\hrule\vspace{\baselineskip}
					\copyright~2025 IEEE. Personal use of this material is permitted. Permission from IEEE must be obtained for all other uses, in any current or future media, including reprinting/republishing this material for advertising or promotional purposes, creating new collective works, for resale or redistribution to servers or lists, or reuse of any copyrighted component of this work in other works.} 
	}\end{spacing}}}
}

\begin{document}
\maketitle

\begin{abstract}
Unsupervised multi-object discovery (MOD) aims to detect and localize distinct object instances in visual scenes without any form of human supervision.
Recent approaches leverage object-centric learning (OCL) and motion cues from video to identify individual objects. 
However, these approaches use supervision to generate pseudo labels to train the OCL model. 
We address this limitation with \MethodName---\textbf{M}otion-\textbf{R}efined \textbf{DINOSAUR}---a minimalistic \emph{unsupervised} approach that extends the self-supervised pre-trained OCL model, DINOSAUR~\cite{Seitzer:2023:BGR}, to the task of unsupervised multi-object discovery.
We generate high-quality unsupervised pseudo labels by retrieving video frames without camera motion for which we perform motion segmentation of unsupervised optical flow. We refine DINOSAUR's slot representations using these pseudo labels and train a slot deactivation module to assign slots to foreground and background. 
Despite its conceptual simplicity, \MethodName achieves strong multi-object discovery results on the TRI-PD and KITTI datasets, outperforming the previous state of the art despite being \emph{fully unsupervised}.
\end{abstract}
 
\thispagestyle{firststyle}
\begin{figure}[t]
    \centering
    \input{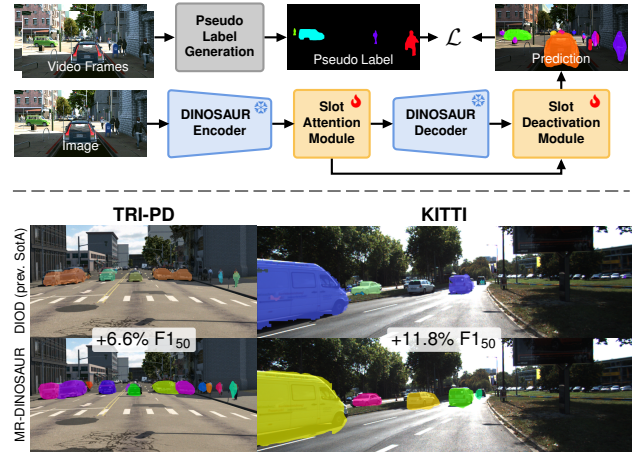}
    \vspace{-1em}
    \caption{\textbf{Overview and results of our unsupervised multi-object discovery approach \MethodName.} We propose a minimalist approach to generate instance pseudo-labels from motion for refining slot representations of the DINOSAUR model. Further, we train our proposed slot deactivation module to distinguish foreground from background slots \textit{(top)}. \MethodName outperforms the previous SotA approach DIOD, indicated by the gains in F1\textsubscript{50} \textit{(bottom)}.}
    \vspace{-2em}
    \label{fig:teaser}
\end{figure}
\section{Introduction\label{sec:introduction}}

Achieving an instance-level understanding of complex scenes is a long-standing challenge in computer vision, with broad applications in autonomous driving, robotics, and medical image analysis~\cite[see][for an overview]{Minaee:2022:ISU, Zhou:2024:ISF}. Approaching instance-level scene understanding via supervised learning requires pixel-wise ground-truth labels \cite{He:2020:MRC, Bolya:2019:YRI, Cai:2019:CRC, Kirillov:2023:SAM}, yet obtaining large amounts of manual pixel-wise annotation is highly resource intensive. For instance, annotating a single \SI{2}{MP} image of a natural scene can take even a trained human annotator up to multiple hours~\cite{Cordts:2016:CDS, Brodermann:2024:MTM}. Approaching instance-level scene understanding using unsupervised learning aims to overcome the challenges associated with human annotations. In our work, we specifically aim to detect and localize distinct object instances in visual scenes without \emph{any} form of human supervision. Hereby, objects are defined as entities capable of moving.

Object-centric learning (OCL) aims at compositional scene understanding by decomposing images into localized object representations using Slot Attention~\cite{Locatello:2020:OLW}.
Trained using self-supervised objectives, recent slot attention-based methods \cite{Seitzer:2023:BGR,Kakogeorgiou:2024:STW,Wu:2023:SOG} have been scaled to work on real-world images.
Though slot-attention methods are able to successfully decompose a scene, using them for instance-level scene understanding remains challenging for two key reasons: \emph{First}, slot-attention approaches specify a fixed number of slots and divide each image into an equal number of segments, including the background. 
Due to using the fixed number of segments, they suffer from poor segmentation of objects, such as over- and under-segmentation~\cite{Zimmermann:2023:SSO}. \emph{Second}, identifying which slot is foreground and background is not trivial.

Bao~\etal~\cite{Bao:2022:DOT} were one of the first to apply slot-attention mechanisms successfully to instance-level scene understanding, which is commonly referred to as multi-object discovery~\cite{Villa:2024:UOD} in the slot-attention literature.
We, henceforth, adopt this terminology. 
Bao \etal \cite{Bao:2022:DOT} relies on synthetic video data~\cite{Dave:2019:TSA} and supervised motion segmentation masks for learning foreground and background slots. Subsequent approaches built on \cite{Bao:2022:DOT} by improving the architecture~\cite{Bao:2023:ODF}, explicitly modeling the background segmentation~\cite{Kara:2024:TBA}, or mitigating noise in motion-based pseudo-labels via self-distillation~\cite{Kara:2024:DSD}. Yet, these methods still entail significant limitations, such as requiring synthetic pre-training to perform well on real-world data and utilizing supervised pipelines for pseudo-label generation.

In this work, we pursue a highly minimalist take on unsupervised multi-object discovery.
We propose \MethodName---\textbf{M}otion \textbf{R}efined \textbf{DINOSAUR}---transferring DINOSAUR \cite{Seitzer:2023:BGR}, a scalable object-centric learning approach, to the task of multi-object discovery (\cf \cref{fig:teaser}). 
On its own, DINOSAUR, like other slot-attention methods, suffers from imprecise object masks and is unable to classify slots as either foreground or background.
We address this by refining the slot representations and learning to distinguish foreground from background using a straightforward and strictly unsupervised motion guidance. By using video frames that entail no camera motion, we can cluster unsupervised optical flow to obtain high-quality instance pseudo-labels as flow is only induced by moving objects. 
Using these pseudo-labels, we propose a novel two-stage training scheme, and slot deactivation module, enabling \MethodName to outperform existing methods for the task of multi-object discovery. 

Specifically, we make the following contributions: \emph{(i)}~We obtain high-precision unsupervised instance pseudo-labels of real-world scenes by retrieving quasi-static frames (\ie., frames with no camera motion) and straightforward motion clustering. \emph{(ii)}~We effectively extend DINOSAUR to the task of multi-object discovery by refining DINOSAUR's predictions and adding a simple, yet effective slot deactivation module. \emph{(iii)}~Despite being minimalistic and fully unsupervised, \MethodName yields state-of-the-art unsupervised multi-object discovery accuracy on KITTI and TRI-PD.

\section{Related Work\label{sec:related_work}}

\inparagraph{Object-centric learning (OCL)} aims to decompose a complex visual scene into a set of semantically meaningful object representations. These representations can be used for compositional generation and downstream tasks such as property prediction \cite{Locatello:2020:OLW}, compositional generation \cite{Akan:2025:SGA,Jiang:2023:OSD}, and building world models \cite{Wu:2023:SUV}. Initial work \cite{Burgess:2019:MUS, Greff:2019:MRL, Elsami:2016:AIR, Li:2020:LOR, LIN:2020:IGI,Lin:2020:SUO,Jiang:2019:SGW} in OCL employed sequential architectures to decompose the input scene, but is hard to scale to complex scenes and imposes an unnatural ordering on objects. To overcome these issues, \cite{Engelcke:2020:GSI,Engelcke:2021:GIU} proposed using stick-breaking priors, while \cite{Locatello:2020:OLW} introduced a slot-attention (SA) mechanism, which uses the standard dot-product attention \cite{Vaswani:2017:AAN} and an iterative refinement scheme. 
Its ease of use and the scalability of dot-product attention have led to the widespread adoption of slot attention for OCL \cite{Kipf:2021:COL, Elsayed:2022:STE, Bao:2022:DOT, Seitzer:2023:BGR}.

\inparagraph{Slot-attention methods for real-world scenes.} 
Until recently, SA methods were trained in the pixel space, which limited their applicability to synthetic datasets \cite{Johnson:2017:CDD,Groth:2018:SLV,Karazija:2021:CTA}.
DINOSAUR~\cite{Seitzer:2023:BGR} scaled SA by leveraging features from self-supervised architectures such as DINO~\cite{Caron:2021:EPS} and learning the slots/object-centric representation by reconstructing these semantic features instead of pixels. SPOT~\cite{Kakogeorgiou:2024:STW} improves DINOSAUR by utilizing student-teacher training. Parallel to these approaches, \cite{Jiang:2019:SGW,Wu:2023:SOG,Singh:2024:GLS,Akan:2025:SGA} have used a diffusion-based decoder for scaling slot attention to real-world scenes. These methods focus on image reconstruction rather than scene decomposition. 
Despite the successes of DINOSAUR~\cite{Seitzer:2023:BGR}, existing SA methods need to specify the number of objects (slots) a-priori for each dataset. This limits these methods, leading to issues of over-segmentation and under-segmentation \cite{Zimmermann:2023:SSO}. We address these issues by refining the slot representation using guidance from unsupervised motion masks and learning to classify each slot into foreground or background.

\inparagraph{Guided multi-object discovery.} 
Unsupervised object discovery for images is an ill-posed problem in the absence of a precise definition of an object. Utilizing video data alleviates this problem, as the principle of common fate~\cite{Koffka:1922:PIGerception} can be used to define what an object is.
Most existing methods utilize additional signals such as depth~\cite{Elsayed:2022:STE}, optical flow~\cite{karazija:2022:UMS,Kipf:2021:COL}, and motion segmentation \cite{Bao:2023:ODF}. However, relying on motion masks makes these methods over-segment the background and unable to segment static objects. BMOD~\cite{Kara:2024:TBA} addressed the issue of over-segmenting the background by introducing and explicitly learning a background slot, guided using masks of moving objects. Another issue in using motion masks for guidance is that these masks are often noisy, making the guidance signal poor; DIOD~\cite{Kara:2024:DSD} tackles this issue and extends BMOD with a self-distillation loss that better handles the noisy guidance masks. 
Compared to earlier works, we simply refine the DINOSAUR model with masks obtained from our novel motion-based pseudo-labeling. Once we refine the encoder model, we train a simple MLP to distinguish between foreground and background slots. 

\begin{figure*}[t]
\centering
\input{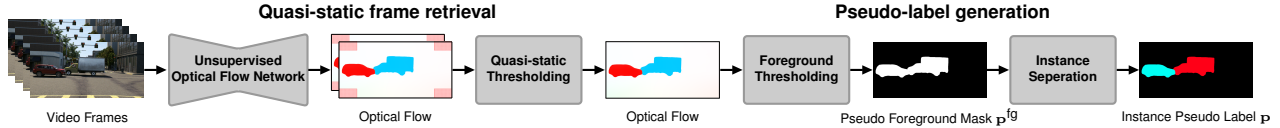}
\vspace{-1.25em}
\caption{\textbf{Pseudo-label generation.} We derive pseudo-instance labels in two stages. First, we perform an unsupervised retrieval of quasi-static frames, \ie those without camera motion, identified by SMURF \cite{Stone:2021:STM} optical flow being small in most of the image corners. 
Given frames with quasi-static camera motion, we perform simplistic motion segmentation, where the foreground mask for moving objects arises by applying a threshold. Connected components and HDBSCAN clustering leads to our instance pseudo label for quasi-static frames.} 
\vspace{-0.5em}
\label{fig:pseudo_label_generation}
\end{figure*}
\inparagraph{Optical flow-based motion segmentation} aims to localize moving objects in imagery by exploiting optical flow~\cite{Anthwal:2019:SMS, Zappella:2008:MSS}. While some approaches only estimate a single foreground motion mask, others obtain object-wise masks~\cite{Zhang:2001:MSS}. Early approaches employed flow discontinuities~\cite{Enkelmann:1991:OBS, William:1985:DCA, Schunck:1989:IFS}, motion trajectories~\cite{Brox:2010:OSG, Ochs:2012:HOM}, or probabilistic motion models~\cite{Torr:1998:GME, Black:2000:PDT, Sommer:2022:SCS}. Recent approaches learned to estimate motion segments using synthetic data with ground-truth masks~\cite{Dave:2019:TSA}, supervised foundation models~\cite{Xie:2024:SAM}, or self-supervised learning~\cite{Safadoust:2023:MOD, Choudhury:2022:GWM}. We propose a simple yet precise motion segmentation approach by first retrieving video frames without camera motion.
Fulfilling the static background assumption significantly simplifies the motion segmentation task.

\section{\MethodName\label{sec:method}}
\begin{figure*}[t]
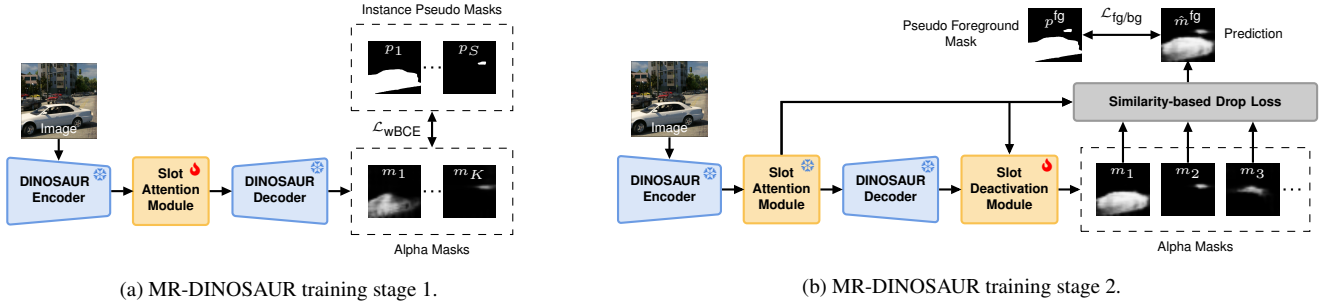

    \centering
    \begin{minipage}{0.42\textwidth}
        \input{artwork/method/Fig2.1}
        \vspace{-0.45em}
        \subcaption{\MethodName training stage 1. \label{fig:main_stage1}}
    \end{minipage}
    \hfill
    \begin{minipage}{0.53\textwidth}
        \input{artwork/method/FIg2.2}
        \vspace{-0.45em}
        \subcaption{\MethodName training stage 2. \label{fig:main_stage2}}
    \end{minipage}
    \vspace{-0.5em}
    \caption{\textbf{\MethodName architecture and training overview.} Stage 1 trains the DINOSAUR slot-attention module using our proposed pseudo-labels. DINOSAUR alpha masks are supervised with pseudo-label instance masks using a weighted binary cross-entropy loss. Stage 2 trains our slot deactivation module using $\mathcal{L}_\text{fg/bg}$ to learn to discriminate between foreground and background slots.\label{fig:main_mrdinosaur}}
\end{figure*}

Our work approaches the problem of multi-object discovery, \ie, segmenting all objects in a scene, where objects are defined as entities capable of moving.
Our novel unsupervised multi-object discovery method is comprised of two stages. First, we generate high-quality unsupervised pseudo labels using motion cues (\cf \cref{sec:pseudo-label}). 
Second, we introduce a pseudo-label-based refinement strategy to extend the OCL method DINOSAUR~\cite{Seitzer:2023:BGR} for multi-object discovery, overcoming its inability to distinguish between foreground and background---that is, between object and non-object regions. An overview of each part is provided in \cref{fig:pseudo_label_generation,fig:main_mrdinosaur}.

\subsection{Pseudo-label generation\label{sec:pseudo-label}}
Our pseudo-label generation comprises two steps: \emph{(1)} we retrieve quasi-static frames, \ie, frames that do not entail camera motion from the training videos, and \emph{(2)} pseudo instance-label generation, for which we employ a simple, yet effective motion segmentation approach on the optical flow of the quasi-static frames. 

\inparagraph{Quasi-static frames retrieval.} 
We retrieve \emph{quasi-static} frames, characterized by minimal camera motion, to generate accurate pseudo instance masks. In these frames, the static background ensures that optical flow predominantly arises from moving objects, allowing for simple clustering to do instance pseudo-labeling.
Given two consecutive video frames $\mathbf{I}_{1},\mathbf{I}_{2}\in\mathbb{R}^{3\times \rm H \times \rm W}$, we compute forward optical flow $\mathbf{f}\in\mathbb{R}^{2\times \rm H \times \rm W}$ using the off-the-shelf \emph{unsupervised} optical flow estimation approach SMURF~\cite{Stone:2021:STM}. To assess the amount of camera motion, we compute the average flow magnitude within each image corner region, denoted as $\|\overline{\mathbf{f}}_{i}\|, \; i\in\{0, 1, 2, 3\}$. 
Since foreground objects typically appear near the image center, we exploit this scene structure. We label frames as quasi-static if at least three of the four corner patches have average flow magnitudes below $\tau_{\text{static}}$. In practice, these corner patches each cover 15\% of the image height and width (Fig.~\ref{fig:pseudo_label_generation}).
We find this simple approach to work surprisingly well (\eg, 99\% Accuracy on KITTI; \cf \cref{tab:pseudo_label_generalization}).

\inparagraph{Pseudo-label generation.} Equipped with quasi-static frames and their optical flow, we aim to obtain instance-wise pseudo masks. We propose a simple pseudo-label generation, depicted in \cref{fig:pseudo_label_generation}.
We first obtain a foreground mask and extract connected components. Subsequently, connected components that potentially comprise multiple objects are partitioned.

First, as quasi-static frames exhibit no camera motion, we simply derive a foreground mask by thresholding the estimated optical flow $\mathbf{f}$. In particular, we threshold the optical flow magnitude $\|f_{:, h, w}\|>\tau_{\text{fg}}$, obtaining a binary foreground mask $\mathbf{p}^{\text{fg}}\in\{0, 1\}^{\rm H \times \rm W}$. To detect individual instances within the foreground mask $\mathbf{p}^{\text{fg}}$, we first extract the connected components~\cite{Di:1999:CCL}, \ie, all spatially disjoint segments. Still, connected components can include one or multiple objects due to overlapping objects. To separate multiple instances within a single connected component, we leverage the common fate principle---pixels with similar motion likely belong to the same object \cite{Wertheimer:1912:ESU}.

To identify connected components that potentially encapsulate multiple objects, we utilize the spatial derivatives of the estimated optical flow. Motion discontinuities signaled by large derivatives suggest that a connected component is composed of multiple objects. In particular, we threshold the norm of the gradient magnitudes $\|(\nabla f_{1, h, w}, \nabla f_{2, h, w})\| > \tau_{\nabla}$ for all points inside each connected component (\ie, ignoring outside contours). If a connected component contains such a sudden change, we will further divide the segment, utilizing HDBSCAN clustering \cite{Campello:2013:DBC}. For each pixel in the connected component, we use the flow magnitude, flow angle, and its pixel position as an input feature for clustering. Our pseudo-label generation produces a variable number $\rm S$ of high-quality instance pseudo masks $\mathbf{p}\in\{0, 1\}^{\rm S \times \rm H \times \rm W}$ per quasi-static frame.

\inparagraph{Discussion.} While there are more advanced motion segmentation approaches available~\cite{Sommer:2022:SCS, Hahn:2025:UPS, Safadoust:2023:MOD, Black:2000:PDT, Xie:2024:SAM, Cremers:2005:MSE, Dave:2019:TSA}, we aim for simplicity and rely on an unsupervised optical flow-based motion clustering approach for pseudo-label generation. Our approach is conceptually simple, captures non-rigid motions, and produces a \emph{variable} number of pseudo-instance masks, accurately discovering foreground objects, albeit only for the subset of quasi-static frames.
\subsection{Refining DINOSAUR}
\label{sec:refining_dinosaur}
We aim to refine and extend DINOSAUR~\cite{Seitzer:2023:BGR}, an established unsupervised object-centric representation learning approach, to perform unsupervised multi-object discovery using our pseudo-instance masks. 
DINOSAUR decomposes an image into a set of regions, each binding to individual objects or distinct scene components, including the background.
While our pseudo-instance masks only capture objects that are actually moving, they provide an informative cue to differentiate between foreground objects and background. To this end, we propose a novel two-stage training scheme. The first stage employs our pseudo-instance masks to enhance the accuracy of the predicted DINOSAUR slot masks. The second stage relies on a novel slot-deactivation module, which enables DINOSAUR to discriminate between foreground and background. Our two-stage training is illustrated in \cref{fig:main_mrdinosaur}. We refer to the resulting model as motion-refined DINOSAUR, or MR-DINOSAUR for short.

\inparagraph{{DINOSAUR}} \cite{Seitzer:2023:BGR} learns a set of representations, decomposing an image into localized and semantically coherent regions. It encodes images with DINO~\cite{Caron:2021:EPS, Oquab:2024:DLR}, followed by a slot-attention mechanism~\cite{Locatello:2020:OLW} to group the features into $\rm K$ slot vectors $\mathbf{z} \in \mathbb{R}^{\rm K \times \rm D_{\text{slots}}}$ via an iterative attention mechanism.
Next, the slot representations are decoded utilizing an MLP. While a transformer decoder can be used as well, Seitzer~\etal~\cite{Seitzer:2023:BGR} found that this leads to slots binding more to semantic representations than objects. We hence use an MLP decoder. Consecutively, each slot representation is independently broadcast onto the image patches,
and a learned positional encoding is added to each token. Each token is processed by the same MLP, yielding a feature reconstruction $\hat{\mathbf{y}}_{k}$ and map $\bm{\alpha}_{k}$ that indicates the region attended to by each slot. The final feature reconstruction is given by
\begin{equation}
\mathbf{y} = \sum_{k=1}^\text{K} \hat{\mathbf{y}}_{k} \odot \mathbf{m}_{k}, \quad \mathbf{m}_{k} = \operatorname*{softmax}\bm{\alpha}_{k}, 
\end{equation}
where $\odot$ denotes the element-wise multiplication.
The training objective is to reconstruct the DINO features with $\mathbf{y}$. In the following, we refer to the softmax masks $\mathbf{m}\in\mathbb{R}^{\rm K \times \rm H \times \rm W}$ as alpha masks.
Our method builds upon a ``vanilla'' DINOSAUR model, trained on the respective dataset as proposed by \cite{Seitzer:2023:BGR}, which is then refined and extended to the task of multi-object discovery by our method.

\inparagraph{Stage 1: Taming the DINOSAUR.} 
Equipped with the pseudo-instance labels from \cref{sec:pseudo-label}, we aim to refine {DINOSAUR's} slot representations to better bind to objects. To that end, training stage 1 fine-tunes the predicted alpha masks  $\mathbf{m}\in\mathbb{R}^{\rm K \times \rm H \times \rm W}$ of the $\rm K$ slots using our pseudo labels $\mathbf{p}$. In particular, we match the alpha masks to the individual masks of the pseudo label using Hungarian matching~\cite{Kuhn:1955:HUN}. We apply a weighted binary cross-entropy loss $\mathcal{L}_{\text{wBCE}}$ using the matched masks. In particular, $\mathcal{L}_{\text{wBCE}}$ between the $\rm S\leq \rm K$ matched predictions $\tilde{\mathbf{m}}_{s}$ and the respective pseudo instance mask $\mathbf{p}_{s}$ for the slot $s$ is defined as:
\begin{equation}
    \begin{aligned}
    \mathcal{L}_{\text{wBCE}}&\bigl({\tilde{\mathbf{m}}}_{s}, \mathbf{p}_{s}\bigr)=\frac{1}{\rm H \cdot \rm W}\sum_{h,w}\Bigl[\bigl(2 - r_{s}\bigr)\mathbf{p}_{s}\\[-2pt]
    &\log\tilde{\mathbf{m}}_{s} + \bigl(1 - \mathbf{p}_{s}\bigr)\log\bigl(1 - \tilde{\mathbf{m}}_{s}\bigr)\Bigr], 
    \end{aligned}
\end{equation}
with weight $r_{s}=\frac{1}{\rm H \cdot \rm W}\sum_{h,w}\mathbf{p}_{s}$, which aids in detecting small objects.
Specifically, if $r_s\to 0$ as the pseudo instance mask $\mathbf{p}_{s}$ only includes a few pixels, then $\bigl(2 - r_{s}\bigr)\to2$, up‐weighting small‐object masks in the loss.
We train the slot-attention module while freezing the encoder and decoder. 
This first training stage already leads to improved slot representations, localizing objects (\cf \cref{sec:experiments}).

\inparagraph{Stage 2: Letting the DINOSAUR see objects.}
While DINOSAUR decomposes the entire image into slots, we extend it to perform unsupervised multi-object discovery by learning to discriminate slots that encode foreground objects from slots encoding background.
Given the refined DINOSAUR model after stage 1 training, we now aim to classify slots into foreground and background using our pseudo labels from \cref{sec:pseudo-label} as the training signal. We strive for a minimalist approach and introduce a slot deactivation module $\phi_d$, which predicts a weight vector $\boldsymbol{\lambda} \in \mathbb{R}^{\rm K}$ from the slot representations $\mathbf{z}$.
The slot-deactivation module is an MLP with a final sigmoid layer predicting a weight $\boldsymbol{\lambda}$ close to 0 for background slots and close to 1 for foreground slots:
\begin{equation}
    \boldsymbol{\lambda} = \phi_d(\mathbf{z}) ,
\end{equation}
where $\lambda_{k}$ is the predicted probability of slot $k$ being classified as foreground. For obtaining the foreground prediction $\mathbf{m}^\text{fg}$ of the model, the alpha masks $\mathbf{m}_{k}$ are simply weighted by the predicted $\lambda_{k}$.
During inference, $\boldsymbol{\lambda}$ is binarized, \ie, if $\lambda_{k} > 0.5$, the corresponding slot $k$ is kept; otherwise, the slot $k$ is deactivated, meaning its alpha mask $\mathbf{m}_k$ is weighted with zero.

\inparagraph{Drop-loss.} In training stage 2, we train the proposed slot-deactivation module while keeping the entire DINOSAUR model frozen (\cref{fig:main_stage2}).
The training is guided by our foreground pseudo-label mask $\mathbf{p}^\text{fg}$ (\cf \cref{sec:pseudo-label}). However, we obtain pseudo-instance masks only for dynamic objects in a scene when captured by a quasi-static camera. This can lead to inconsistent training signals for the foreground/background separation if an object of the same category is static in another frame (\eg, driving \vs parked cars) and, therefore, the corresponding slot representation would be considered once as foreground and once as background. 

To allow our model to discover objects beyond the dynamic ones captured in our pseudo labels, we carefully select the relevant alpha mask predictions $\hat{\mathbf{m}}^\text{fg}$ to be considered in the loss computation and ignore predictions in the loss that might be correct (static objects) but are not part of the pseudo label (dynamic objects). 
Inspired by \cite{Wang:2023:CLU}, we propose a similarity-based drop-loss mechanism, where we ignore a slot in our loss term if its alpha mask is not matched to any pseudo-instance mask, but its slot representation yields high cosine similarity to the representation of one of the matched slots.
Specifically, after matching, we have $S$ matched alpha mask predictions $\tilde{\mathbf{m}}_{s}$ and slot representations $\tilde{\mathbf{z}}_{s}$, as well as $U = K - S$ unmatched alpha mask predictions $\bar{\mathbf{m}}_{u}$ and slot representations $\bar{\mathbf{z}}_{u}$.
We calculate the cosine similarity of each unmatched slot representation to all matched slot representations as $\mathbf{c}_u = \operatorname{cos-sim}(\bar{\mathbf{z}}_{u}, \tilde{\mathbf{z}}_s)$.
When applying the similarity-based drop-loss mechanism, the foreground prediction  $\hat{\mathbf{m}}^\text{fg}$ considered in the loss arises as follows. 
We consider all matched slots (dynamic objects) and all unmatched slots that are dissimilar to the matched slots (background):
\begin{equation}
    \hat{\mathbf{m}}^\text{fg} = \sum_{s}\lambda_s \mathbf{m}_{s} + \sum_{u}\mathbbm{1}\bigl(\max \mathbf{c}_u \leq \tau_{\text{drop}}\bigr) \lambda_u \mathbf{m}_{u},
\end{equation}
where $\tau_{\text{drop}}$ refers to the slot similarity threshold (\eg, $\tau_{\text{drop}} = 0.99 $).

Given $\hat{\mathbf{m}}^\text{fg}$ and the foreground pseudo-label $\mathbf{p}^\text{fg}$, we employ a foreground-background loss.
This loss combines a negative log-likelihood (NLL) loss to learn the foreground and a background regularization term.
Hereby, we prevent the model from collapsing to the trivial solution of predicting foreground for all pixels. The foreground-background loss is defined as
\begin{equation}
    \mathcal{L}_\text{fg/bg}(\hat{\mathbf{m}}^\text{fg}, \mathbf{p}^\text{fg}) = \mathcal{L}_\text{fg}(\hat{\mathbf{m}}^\text{fg}, \mathbf{p}^\text{fg}) + \mathcal{L}_\text{bg}(\hat{\mathbf{m}}^\text{fg}, \mathbf{p}^\text{fg}),
\end{equation}
where the foreground NLL loss is given as
\begin{equation}
    \mathcal{L}_\text{fg}(\hat{\mathbf{m}}^\text{fg}, \mathbf{p}^\text{fg}) = \frac{1}{\rm H \cdot \rm W}\sum_{h,w} {\mathbf{p}}^\text{fg} \log\hat{\mathbf{m}}^\text{fg}.
\end{equation}
Given the number of background pixels $\rm N^p_\text{bg}$ in the pseudo-label $\mathbf{p}^\text{fg}$, the background regularization component is defined as
\begin{equation}
\mathcal{L}_\text{bg}(\hat{\mathbf{m}}^\text{fg}, \mathbf{p}^\text{fg}) = \frac{r_\text{bg}}{\rm N^p_\text{bg}}\sum_{h,w} \mathbbm{1}\bigl({\mathbf{p}}^\text{fg}=0\bigr)\hat{\mathbf{m}}^\text{fg}.
\end{equation}

\section{Experiments\label{sec:experiments}}

We compare \MethodName against state-of-the-art slot-attention methods for video and image object discovery across multiple datasets, following the common evaluation protocol \cite{Bao:2022:DOT, Bao:2023:ODF, Kara:2024:TBA, Kara:2024:DSD}.
We refer to the supplemental material for additional results. 

\inparagraph{Datasets.} To train and evaluate, we use two datasets: First, TRI-PD~\cite{Bao:2022:DOT}, a synthetic dataset simulating driving scenarios. The training set comprises 924 photorealistic videos. Each video is 10 seconds long and captured at 20 FPS. 51 additional videos from disjoint scenes serve for evaluation.
Second, KITTI~\cite{Geiger:2013:VMR}, a real-world dataset capturing urban driving scenarios. We train on KITTI raw data and evaluate on the 200 images of the instance segmentation subset. We exclude the validation samples from the training data.

\begin{table}[t]
    \centering
    \caption{\textbf{Unsupervised multi-object discovery on TRI-PD} using F1\textsubscript{50}, AP\textsubscript{50}, AR\textsubscript{50}, all-ARI, and fg-ARI (all in \%, $\uparrow$). $*$ denotes methods using DINOv2. \underline{Underlined methods} use supervision. \label{tab:tripd_main}}
    \vspace{-0.5em}
    \sisetup{table-number-alignment=center}
\newcommand{\mytablecolumnwidth}{0.11\linewidth}
\footnotesize
\setlength{\tabcolsep}{1pt}
\renewcommand{\arraystretch}{0.99}
\begin{tabularx}{\columnwidth}{
>{\hspace{-\tabcolsep}\raggedright\columncolor{white}[\tabcolsep][\tabcolsep]}
X>{\centering\arraybackslash}
S[table-format=2.1, table-column-width=\mytablecolumnwidth]
S[table-format=2.1, table-column-width=\mytablecolumnwidth]
S[table-format=2.1, table-column-width=\mytablecolumnwidth]
S[table-format=2.1, table-column-width=\mytablecolumnwidth]
S[table-format=2.1, table-column-width=\mytablecolumnwidth]
S[table-format=2.1, table-column-width=\mytablecolumnwidth]
S[table-format=2.1, table-column-width=\mytablecolumnwidth]@{}}
  \toprule
  {\textbf{Method}} & {\textbf{F1\textsubscript{50}}} & {\textbf{AP\textsubscript{50}}} & {\textbf{AR\textsubscript{50}}} & {\textbf{all-ARI}} & {\textbf{fg-ARI}}\\
  \midrule
  SlotAttention \cite{Locatello:2020:OLW} \scriptsize{\textcolor{tud0c!80}{NeurIPS'20}} & {--} & {--} & {--} & {--} & 10.2 \\ 
  MONet \cite{Burgess:2019:MUS}  \scriptsize{\textcolor{tud0c!80}{arXiv'19}}         & {--} & {--} & {--} & {--} & 11.0 \\ 
  SCALOR \cite{Jiang:2020:GWM} \scriptsize{\textcolor{tud0c!80}{ICLR'20}}          & {--} & {--} & {--} & {--} & 18.6 \\ 
  IODINE \cite{Greff:2019:MRL}  \scriptsize{\textcolor{tud0c!80}{ICML'19}}         & {--} & {--} & {--} & {--} & 9.8  \\ 
  MCG \cite{PontTuset:2015:MCG} \scriptsize{\textcolor{tud0c!80}{TPAMI'15}}         & {--} & {--} & {--} & {--} & 25.1 \\     
  \underline{Bao \etal}~\cite{Bao:2022:DOT} \scriptsize{\textcolor{tud0c!80}{CVPR'22}} & 12.2 & {--} & {--} & 6.3 & 50.9 \\ 
  \underline{MoTok}~\cite{Bao:2023:ODF}  \scriptsize{\textcolor{tud0c!80}{CVPR'23}}   & 12.6 & {--} & {--} & 4.7 & 55.1 \\ 
  \underline{BMOD}~\cite{Kara:2024:TBA} \scriptsize{\textcolor{tud0c!80}{WACV'24}}    & 14.4 & {--} & {--} & 28.6 & 53.9 \\ 
  \underline{BMOD}$^*$~\cite{Kara:2024:TBA} \scriptsize{\textcolor{tud0c!80}{WACV'24}} & 16.3 & {--} & {--} & 29.1 & 58.5 \\  
  \underline{DIOD}~\cite{Kara:2024:DSD} \scriptsize{\textcolor{tud0c!80}{CVPR'24}}    & 35.4 & 28.9 & 45.6 & 70.3 & 66.1 \\
  \underline{DIOD}$^*$~\cite{Kara:2024:DSD} \scriptsize{\textcolor{tud0c!80}{CVPR'24}}  & 41.5 & 37.8 & 45.9 & \textbf{74.1} & 69.7 \\  
  \midrule
  DINOSAUR$^*$ \cite{Seitzer:2023:BGR} \scriptsize{\textcolor{tud0c!80}{ICLR'23}}     & 4.9 & 2.8 & 19.8 & 2.2 & 52.0\\ 
  \rowcolor{tud0c!20} MR-DINOSAUR$^*$ \textit{(Ours)} &
    \textbf{48.1} & \textbf{45.1} & \textbf{51.5} & 71.9 & \textbf{74.4} \\
  \bottomrule
\end{tabularx}

\end{table}

\inparagraph{Metrics.\label{sec:metrics}} We evaluate using F1\textsubscript{50}, the harmonic mean of precision AP\textsubscript{50} and recall AR\textsubscript{50} at a mask IoU threshold of $50\,\%$. F1\textsubscript{50} inherently normalizes object sizes and effectively penalizes background over-segmentation by treating each spurious background segment as a false positive. In addition, we use fg-ARI, a standard metric in object discovery that quantifies the similarity between predicted and ground-truth clusterings by computing the ARI only over foreground regions. 
In addition, we use all-ARI~\cite{Kara:2024:TBA}, which incorporates background regions in the ARI computation. However, the pixel-wise nature of ARI metrics inherently favors the accurate segmentation of larger objects, as correctly clustering many pixels disproportionately boosts the score.
Note that recent papers have raised concerns regarding the reliability of the fg-ARI metric \cite{Engelcke:2020:GSI, Monnier:2021:ULI, Karazija:2021:CTA, Seitzer:2023:BGR, Wu:2023:SOG, Kakogeorgiou:2024:STW}. In particular, fg-ARI has been shown to be sensitive to segmentation biases, potentially favoring either over-segmentation \cite{Engelcke:2020:GSI, Monnier:2021:ULI} or under-segmentation. In addition, excluding background pixels from the evaluation may not fully capture object segmentation performance \cite{Karazija:2021:CTA, Monnier:2021:ULI}. ARI results must thus be considered with caution.

\inparagraph{Baseline.} We utilize DINOSAUR~\cite{Seitzer:2023:BGR} and follow the training and inference setup proposed by Seitzer \etal \cite{Seitzer:2023:BGR}.
Hereby, inference on non-square images is performed using a non-overlapping sliding window approach with square image crops. To address objects spanning across multiple windows being segmented by multiple masks, we incorporate the slot merger mechanism \cite{Aydemir:2023:SOC} for evaluation. 
Specifically, masks located at the borders of the sliding window crops are merged utilizing agglomerative clustering based on the cosine similarity of the slot representations.

\inparagraph{Implementation details.} We use PyTorch~\cite{Paszke:2019:PAT} and build upon the code of DINOSAUR~\cite{Seitzer:2023:BGR}, SMURF~\cite{Stone:2021:STM}, and DIOD~\cite{Kara:2024:DSD}.
For pseudo-label generation, we apply a quasi-static frame retrieval threshold of $\tau_{\text{static}} = 0.5$ for TRI-PD and $\tau_{\text{static}} = 1.7$ for KITTI. We use a foreground mask threshold of $\tau_{\text{fg}} = 2.5$, and an inner mask flow-gradient threshold of $\tau_{\nabla} = 20$.
We utilize the DINOSAUR framework to train the baseline models with DINOv2~\cite{Oquab:2024:DLR} encoder using 60 slots in total per image. Specifically, we use $2 \times 30$ slots for TRI-PD and $4 \times 15$ slots for KITTI.
The background loss weight $r_\text{bg}$ is set to 0.2, the similarity-based drop loss uses $\tau_\text{drop}=0.99$.
The slot-deactivation module is a four-layer MLP with a hidden dimension of 2048 and a sigmoid activation for the last layer.
Following previous work \cite{Bao:2022:DOT, Bao:2023:ODF, Kara:2024:TBA, Kara:2024:DSD}, we resize and crop images to a resolution of $980 \times 490$ for TRI-PD and $1260 \times 378$ for KITTI for both pseudo-labeling and training.
Training is performed on two non-overlapping square crops of size $490$ for TRI-PD and four overlapping crops of size $378$ for KITTI.
Stage 1 trains for 15 epochs using a learning rate of $4\mathrm{e}{-6}$. In stage 2, training is performed for one epoch using a learning rate of $4\mathrm{e}{-5}$.
We use a batch size of 8 and optimize with Adam~\cite{Kingma:2015:AAM}.
All experiments use a \emph{single} NVIDIA A6000 Ada GPU. We refer to the supplement for more details.

\begin{table}[t]
    \centering
    \caption{\textbf{Unsupervised multi-object discovery on KITTI} using F1\textsubscript{50}, AP\textsubscript{50}, AR\textsubscript{50}, all-ARI, and fg-ARI (all in \%, $\uparrow$). $*$ denotes methods using DINOv2, $\dagger$ methods pre-trained on synthetic TRI-PD data, and $\ddagger$ re-training on KITTI only, for a fair comparison. 
    \underline{Underlined methods} use supervision.
    \label{tab:kitti_main}}
    \vspace{-0.5em}
    \sisetup{table-number-alignment=center}
\newcommand{\mytablecolumnwidth}{0.11\linewidth}
\footnotesize
\setlength{\tabcolsep}{1pt}
\renewcommand{\arraystretch}{0.99}

\begin{tabularx}{\columnwidth}{
  >{\hspace{-\tabcolsep}\raggedright\columncolor{white}[\tabcolsep][\tabcolsep]}X
  >{\centering\arraybackslash}
  S[table-format=2.1, table-column-width=\mytablecolumnwidth]
  S[table-format=2.1, table-column-width=\mytablecolumnwidth]
  S[table-format=2.1, table-column-width=\mytablecolumnwidth]
  S[table-format=2.1, table-column-width=\mytablecolumnwidth]
  S[table-format=2.1, table-column-width=\mytablecolumnwidth]}
  \toprule
  \textbf{Method} &
  \textbf{F1\textsubscript{50}} &
  \textbf{AP\textsubscript{50}} &
  \textbf{AR\textsubscript{50}} &
  \textbf{all-ARI} &
  \textbf{fg-ARI} \\
  \midrule
  SlotAttention \cite{Locatello:2020:OLW} \scriptsize{\textcolor{tud0c!80}{NeurIPS'20}} & {--} & {--} & {--} & {--} & 13.8 \\
  MONet \cite{Burgess:2019:MUS} \scriptsize{\textcolor{tud0c!80}{arXiv'19}}         & {--} & {--} & {--} & {--} & 14.9 \\
  SCALOR \cite{Jiang:2020:GWM} \scriptsize{\textcolor{tud0c!80}{ICLR'20}}          & {--} & {--} & {--} & {--} & 21.1 \\
  IODINE \cite{Greff:2019:MRL} \scriptsize{\textcolor{tud0c!80}{ICML'19}}          & {--} & {--} & {--} & {--} & 14.4 \\
  MCG \cite{PontTuset:2015:MCG} \scriptsize{\textcolor{tud0c!80}{TPAMI'15}}         & {--} & {--} & {--} & {--} & 40.9 \\
  STEVE \cite{Singh:2022:SUO}  \scriptsize{\textcolor{tud0c!80}{NeurIPS'22}}          & {--} & {--} & {--} & {--} & 11.9 \\
  \underline{SAVI} \cite{Kipf:2021:COL} \scriptsize{\textcolor{tud0c!80}{ICLR'21}} & {--} & {--} & {--} & {--} & 20.0 \\
  \underline{PPMP} \cite{karazija:2022:UMS} \scriptsize{\textcolor{tud0c!80}{NeurIPS'22}} & {--} & {--} & {--} & {--} & 51.9 \\
  \underline{SAVI++} \cite{Elsayed:2022:STE} \scriptsize{\textcolor{tud0c!80}{NeurIPS'22}} & {--} & {--} & {--} & {--} & 23.9 \\
  \underline{Bao \textit{et al.}}$^\dagger$ \cite{Bao:2022:DOT} \scriptsize{\textcolor{tud0c!80}{CVPR'22}} & 8.8 & {--} & {--} & 4.2 & 47.1 \\
  \underline{MoTok}$^\dagger$ \cite{Bao:2023:ODF} \scriptsize{\textcolor{tud0c!80}{CVPR'23}} & 8.2 & {--} & {--} & 2.1 & 64.4 \\
  \underline{BMOD}$^\dagger$ \cite{Kara:2024:TBA} \scriptsize{\textcolor{tud0c!80}{WACV'24}} & 9.3 & {--} & {--} & 17.8 & 54.7 \\
  \underline{BMOD}$^{*\dagger}$ \cite{Kara:2024:TBA} \scriptsize{\textcolor{tud0c!80}{WACV'24}} & 10.9 & {--} & {--} & 21.7 & 60.8 \\
  \underline{DIOD}$^\dagger$ \cite{Kara:2024:DSD} \scriptsize{\textcolor{tud0c!80}{CVPR'24}} & 18.0 & 17.6 & 18.4 & 61.6 & 73.5 \\
  \underline{DIOD}$^{*\dagger}$ \cite{Kara:2024:DSD}  \scriptsize{\textcolor{tud0c!80}{CVPR'24}} & 23.2 & 26.3 & 20.8 & \bfseries 81.6 & 72.3 \\
  \underline{DIOD}$^{*\ddagger}$ \cite{Kara:2024:DSD}  \scriptsize{\textcolor{tud0c!80}{CVPR'24}} & 14.1 & 14.4 & 13.8 & 27.7 & 52.6 \\
  \midrule
  DINOSAUR$^*$ \cite{Seitzer:2023:BGR} \scriptsize{\textcolor{tud0c!80}{ICLR'23}} & 5.7 & 3.6 & 13.7 & 1.1 & 72.0 \\
  \rowcolor{tud0c!20}
    MR-DINOSAUR$^*$ \textit{(Ours)} & \bfseries 35.0 & \bfseries 59.7 & \bfseries 24.7 & 74.9 & \bfseries 74.1 \\
  \bottomrule
\end{tabularx}

\end{table}
\subsection{Comparison to the state of the art}

The most competitive methods, following the same task setup as we do, include Bao~\etal~\cite{Bao:2022:DOT}, BMOD~\cite{Kara:2024:TBA}, and DIOD~\cite{Kara:2024:DSD}. Since these methods use the same pseudo-labels~\cite{Bao:2022:DOT} generated with the supervised motion segmentation approach TSAM~\cite{Dave:2019:TSA}, all methods implicitly use some form of supervision and cannot be considered completely unsupervised.
In contrast, our approach \MethodName is \emph{fully} unsupervised.
An additional limitation of prior work when applied to real-world images is the reliance on pre-training with synthetic images.
\begin{table*}[t]
    \renewcommand{\arraystretch}{0.95}
    \caption{\textbf{Analyzing \MethodName} training stage contribution \emph{\subref{tab:analysis_components_smurf}}, drop loss contribution \emph{\subref{tab:analysis_loss}}, and number of available slots \emph{\subref{tab:analysis_numslots_smurf}}. We report F1\textsubscript{50}, all-ARI, fg-ARI (all in \%, $\uparrow$) trained and evaluated on the TRI-PD dataset. \label{tab:analysis_architecture}}
    \vspace{-0.5em}
    \begin{minipage}{0.32\textwidth}
    \subcaption{Training analysis. \label{tab:analysis_components_smurf}}
    \vspace{-0.25em}
    \sisetup{table-number-alignment=center}
\newcommand{\mytablecolumnwidth}{0.17\linewidth}
\footnotesize
\setlength{\tabcolsep}{0.1pt}
\renewcommand{\arraystretch}{0.85}
\begin{tabularx}{\columnwidth}{
>{\hspace{-\tabcolsep}\raggedright\columncolor{white}[\tabcolsep][\tabcolsep]}
X>{\centering\arraybackslash}
S[table-format=2.1, table-column-width=\mytablecolumnwidth]
S[table-format=2.1, table-column-width=\mytablecolumnwidth]
S[table-format=2.1, table-column-width=\mytablecolumnwidth]@{}}
  \toprule
  \,\textbf{Method} & \textbf{F1\textsubscript{50}} & \textbf{all-ARI} & \textbf{fg-ARI} \\
  \midrule
  {\,DINOSAUR$^*$\cite{Seitzer:2023:BGR}} & 4.9 & 2.2 & 52.0 \\ 
  {\,+ Stage 1} & 12.5 & 3.3 & 75.6 \\
  \rowcolor{tud0c!20}{\,+ Stage 2} & 48.1 & 71.9 & 74.4 \\ 
  \midrule
  \,+ Only Stage 2 & 25.9 & 68.8 & 50.9 \\
  \bottomrule
\end{tabularx}

\end{minipage}
\hfill
    \begin{minipage}{0.32\textwidth}
        \subcaption{Drop loss analysis.\label{tab:analysis_loss}}
    \vspace{-0.25em}
    \sisetup{table-number-alignment=center}
\newcommand{\mytablecolumnwidth}{0.17\linewidth}
\footnotesize
\setlength{\tabcolsep}{0.1pt}
\renewcommand{\arraystretch}{1.2}
\begin{tabularx}{\columnwidth}{
>{\hspace{-\tabcolsep}\raggedright\columncolor{white}[\tabcolsep][\tabcolsep]}
X>{\centering\arraybackslash}
S[table-format=2.1, table-column-width=\mytablecolumnwidth]
S[table-format=2.1, table-column-width=\mytablecolumnwidth]
S[table-format=2.1, table-column-width=\mytablecolumnwidth]@{}}
  \toprule
  \,\textbf{Method} & \textbf{F1\textsubscript{50}} & \textbf{all-ARI} & \textbf{fg-ARI} \\
  \midrule 
  {\,DINOSAUR$^*$\cite{Seitzer:2023:BGR}} & 4.9 & 2.2 & 52.0\\
  {\,Ours w/o drop loss} & 46.3 & 70.5 & 73.3 \\
  \rowcolor{tud0c!20}{\,Ours w/ drop loss} & 48.1 & 71.9 & 74.4 \\
  \bottomrule
\end{tabularx}

\end{minipage}
\hfill
    \begin{minipage}{0.32\textwidth}
        \subcaption{Number of slots analysis. \label{tab:analysis_numslots_smurf}}
    \vspace{-0.25em}
    \sisetup{table-number-alignment=center}
\newcommand{\mytablecolumnwidth}{0.17\linewidth}
\footnotesize
\setlength{\tabcolsep}{0.1pt}
\renewcommand{\arraystretch}{1.2}
\begin{tabularx}{\columnwidth}{
>{\hspace{-\tabcolsep}\raggedright\columncolor{white}[\tabcolsep][\tabcolsep]}
X>{\centering\arraybackslash}
S[table-format=2.1, table-column-width=\mytablecolumnwidth]
S[table-format=2.1, table-column-width=\mytablecolumnwidth]
S[table-format=2.1, table-column-width=\mytablecolumnwidth]@{}}
	\toprule
    {\,\textbf{Num. Slots}} & {\textbf{F1\textsubscript{50}}} & {\textbf{all-ARI}} & {\textbf{fg-ARI}} \\
    \midrule 
        {\,40} & 45.9 & 69.2 & 73.9 \\

    \rowcolor{tud0c!20} {\,60} & 48.1 & 71.9& 74.4 \\
    {\,80} & 43.8 & 67.6 & 70.2 \\
	\bottomrule
\end{tabularx}

\end{minipage}
\vspace{-0.5em}
\end{table*}
First, we compare our method against recent work on the synthetic TRI-PD dataset in \cref{tab:tripd_main}. \MethodName outperforms competing approaches across all metrics, except for all-ARI, where we score slightly lower.
In particular, our method improves F1\textsubscript{50} by $6.6\,\%$ points over the state-of-the-art DIOD~\cite{Kara:2024:DSD}.
Regarding the ARI metrics, we would like to refer to \cref{sec:metrics} summarizing the concerns of previous works \cite{Engelcke:2020:GSI, Monnier:2021:ULI, Karazija:2021:CTA, Seitzer:2023:BGR, Wu:2023:SOG, Kakogeorgiou:2024:STW} and shift the focus to the established F1\textsubscript{50}.
While BMOD~\cite{Kara:2024:TBA} and DIOD~\cite{Kara:2024:DSD} incorporate temporal information from multiple frames during both training and inference on TRI-PD, we use video frame pairs solely for pseudo-label generation. Second, \cref{tab:kitti_main}, we compare against prior work on the KITTI dataset.  
\MethodName outperforms the previous approaches across fg-ARI, F1\textsubscript{50}, AP\textsubscript{50}, and AR\textsubscript{50}. We perform slightly worse in terms of the all-ARI metric.
Outperforming the state of the art on the challenging F1\textsubscript{50} metric is a significant result given that \MethodName does not use any form of supervision.
For comparison on equal footing, we trained the current state-of-the-art approach DIOD on the real-world KITTI data only, without pre-training on TRI-PD (\cf \cref{tab:kitti_main}; \underline{DIOD}$^{*\ddagger}$). This leads to a severe drop in DIOD's results despite using the same DINOv2 features as our approach.
\MethodName improves $11.8\,\%$ points over the state-of-the-art DIOD~\cite{Kara:2024:DSD} and $20.9\,\%$ points over DIOD trained only on KITTI in terms of F1\textsubscript{50}.

\subsection{Analyzing \MethodName}

We analyze the individual components of \MethodName through several ablation experiments.

\inparagraph{Architecture analysis.}
In \cref{tab:analysis_architecture}, we analyze the impact of our architectural decisions and components.
We find that both training stages boost multi-object discovery metrics (\cf \cref{tab:analysis_components_smurf}).
Stage 1 refines the slot representations to bind objects more effectively, while stage 2 learns foreground segmentation, which increases the F1 score. Training only stage 2 for foreground-background distinction is possible, but leads to significantly worse results.
In \cref{tab:analysis_loss}, we train \MethodName with and without our proposed similarity-based drop loss. We find that the drop loss enables the slot deactivation module to explore objects beyond the moving instances in the pseudo-labels.
Finally, we examine different numbers of slots in \cref{tab:analysis_numslots_smurf}. While 60 slots yield a slight performance advantage, using 40 or 80 slots also produces good results.

\inparagraph{Pseudo-label analysis.\label{sec:pseudolabel_analysis}}
We evaluate our pseudo-labels both quantitatively and qualitatively. In \cref{tab:analysis_pseudolabels}, we analyze our pseudo-labels and compare them with the TSAM pseudo-labels introduced by Bao~\etal~\cite{Bao:2022:DOT} and adopted in recent works~\cite{Bao:2023:ODF, Kara:2024:TBA, Kara:2024:DSD}. Our pseudo-labels differ fundamentally from TSAM labels. While TSAM provides a pseudo-label for every video frame---which introduces challenges for motion segmentation when the camera is moving---our approach generates pseudo-labels only for quasi-static frames.
We compare TSAM and our pseudo-labels using both unsupervised SMURF~\cite{Stone:2021:STM} flow predictions used across all main experiments, and supervised RAFT~\cite{Teed:2021:RAP} flow predictions. We evaluate the subset of images pseudo-labeled by our approach. Due to differences in motion prediction, the image sets for SMURF and RAFT pseudo-labels differ slightly.
We restrict the evaluation to the $~95\,\%$ shared samples.
Our method outperforms the TSAM labels by a large margin. This is a significant result given that TSAM uses supervised training compared to our minimalistic unsupervised pseudo-labeling approach.
\begin{table}[t]
    \centering
    \caption{\textbf{\MethodName pseudo-label analysis} using F1\textsubscript{50}, all-ARI, fg-ARI (all in \% , $\uparrow$) on the TRI-PD dataset.\label{tab:analysis_pseudolabels}}
    \vspace{-0.5em}
    \sisetup{table-number-alignment=center}
\newcommand{\mytablecolumnwidth}{0.112\linewidth}
\footnotesize
\setlength{\tabcolsep}{1pt}
\renewcommand{\arraystretch}{0.99}
\begin{tabularx}{\columnwidth}{
>{\hspace{-\tabcolsep}\raggedright\columncolor{white}[\tabcolsep][\tabcolsep]}
X>{\centering\arraybackslash}c
S[table-format=2.1, table-column-width=\mytablecolumnwidth]
S[table-format=2.1, table-column-width=\mytablecolumnwidth]
S[table-format=2.1, table-column-width=\mytablecolumnwidth]
S[table-format=2.1, table-column-width=\mytablecolumnwidth]
S[table-format=2.1, table-column-width=\mytablecolumnwidth]@{}}
  \toprule
  {\textbf{Method}} & {\textbf{\# Samples}} & {\textbf{F1\textsubscript{50}}} & {\textbf{all-ARI}} & {\textbf{fg-ARI}} \\
  \midrule 
  {TSAM~\cite{Bao:2022:DOT}} & 93729 & 8.9 & 18.1 & 20.2 \\ 
  \midrule
  {TSAM~\cite{Bao:2022:DOT}} & 12514 & 7.1 & 20.8 & 18.4 \\ 
  \rowcolor{tud0c!20}{\textit{Ours}} & 12514 & 15.4 & 32.9 & 41.5 \\ 
  {\textit{Ours} (w/ RAFT~\cite{Teed:2021:RAP})} & 12514 & 15.1 & 34.1 & 41.1 \\ 
  \bottomrule
\end{tabularx}

\end{table}
Figure~\ref{fig:pseudo_qualitative} shows qualitative results of our pseudo-labels for both KITTI and TRI-PD. We observe that our pseudo-instance masks exhibit high quality and precisely align with the contours of the moving objects.
\begin{figure}[t]           \newcommand{\kittiimgwidth}{0.28}
\newcommand{\trypdcorrection}{0.6038647343}
\newcommand{\tripdimgwidth}{\fpeval{\trypdcorrection * \kittiimgwidth}}
\scriptsize
\sffamily
\setlength{\tabcolsep}{1pt}
\renewcommand{\arraystretch}{0.0}
\begin{tabular}{>{\centering\arraybackslash} m{0.02\textwidth} 
                >{\centering\arraybackslash} m{\tripdimgwidth\textwidth}
                >{\centering\arraybackslash} m{\kittiimgwidth\textwidth} }

\rotatebox[origin=l]{90}{\scriptsize{\hspace{-18.7em}Pseudo Label\hspace{1.5em}Motion\hspace{3.1em}Image}} & {\textbf{TRI-PD}} & {\textbf{KITTI}} \\[2pt]

& \includegraphics[width=\linewidth]{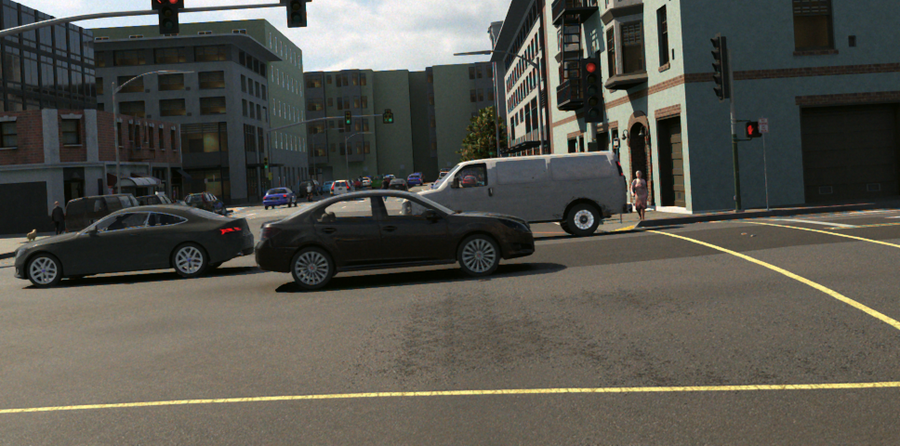}& \includegraphics[width=\linewidth]{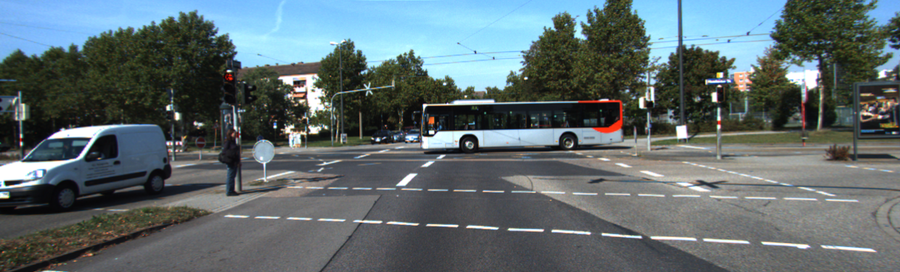}  \\

& \includegraphics[width=\linewidth]{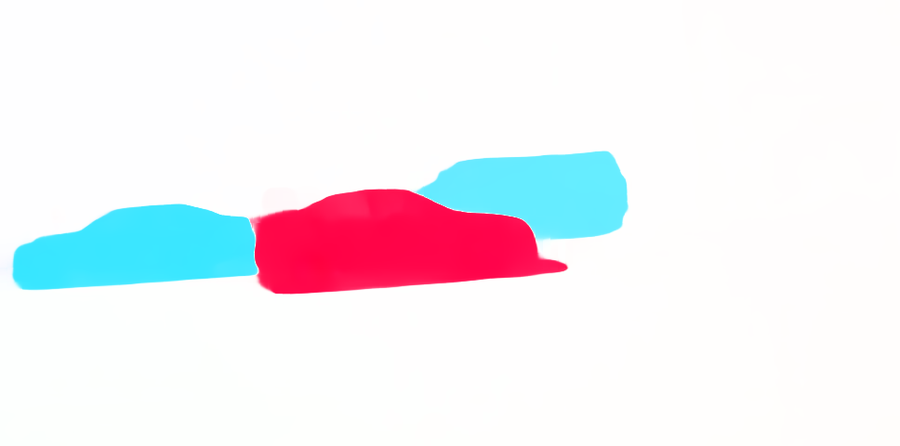}
& \includegraphics[width=\linewidth]{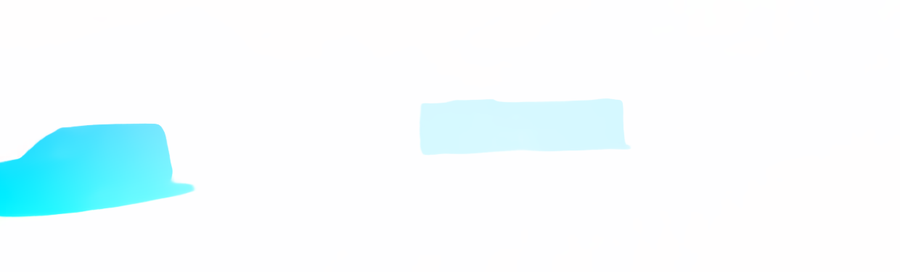} \\

& \includegraphics[width=\linewidth]{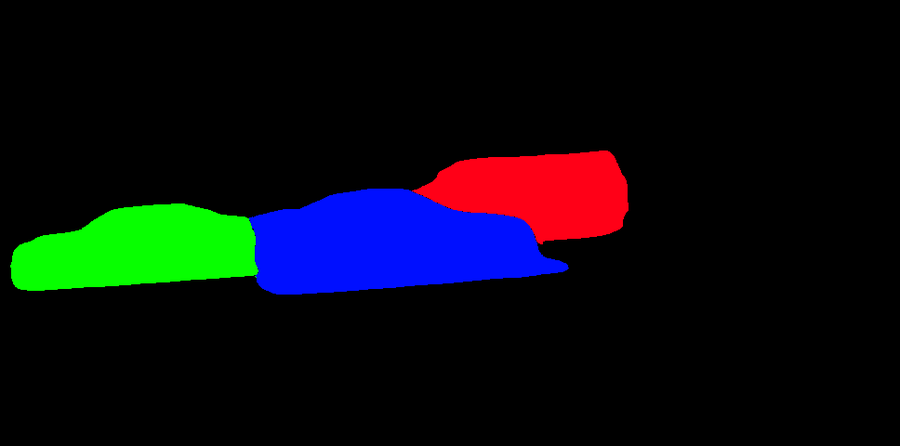}
& \includegraphics[width=\linewidth]{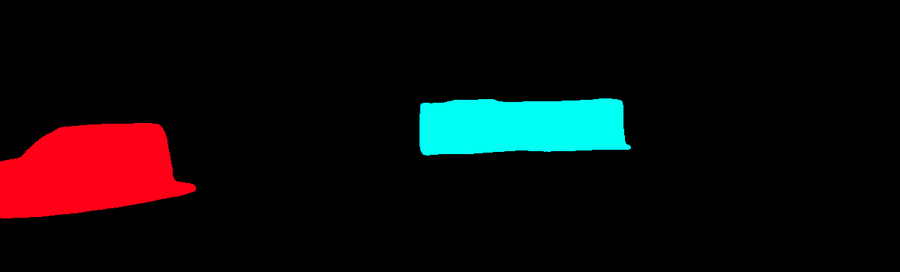}  \\
\end{tabular}

\vspace{-0.5em}
\caption{\textbf{Qualitative examples of our proposed pseudo-labels} on the TRI-PD~\cite{Bao:2022:DOT} and KITTI~\cite{Geiger:2013:VMR} dataset. \label{fig:pseudo_qualitative}}
\end{figure}
\begin{table}[t]
    \centering
    \caption{\textbf{\MethodName with RAFT pseudo-labels.} We report F1\textsubscript{50}, all-ARI, and fg-ARI (all in \%, $\uparrow$) on the TRI-PD dataset. \label{tab:raft_training}}
    \vspace{-0.5em}
    \sisetup{table-number-alignment=center}
\newcommand{\mytablecolumnwidth}{0.112\linewidth}
\footnotesize
\setlength{\tabcolsep}{1pt}
\renewcommand{\arraystretch}{0.99}
\begin{tabularx}{\columnwidth}{
>{\hspace{-\tabcolsep}\raggedright\columncolor{white}[\tabcolsep][\tabcolsep]}
X>{\centering\arraybackslash}c
S[table-format=2.1, table-column-width=\mytablecolumnwidth]
S[table-format=2.1, table-column-width=\mytablecolumnwidth]
S[table-format=2.1, table-column-width=\mytablecolumnwidth]
S[table-format=2.1, table-column-width=\mytablecolumnwidth]
S[table-format=2.1, table-column-width=\mytablecolumnwidth]@{}}
  \toprule
  \textbf{Method} & \textbf{Pseudo Labels} & \textbf{F1\textsubscript{50}} & \textbf{all-ARI} & \textbf{fg-ARI}\\
  \midrule 
  DINOSAUR$^*$\cite{Seitzer:2023:BGR}                & {--}   &  4.9 &  2.2 & 52.0 \\
  \rowcolor{tud0c!20}MR-DINOSAUR$^*$\textit{(Ours)}  & SMURF  & 48.1 & 71.9 & 74.4 \\
  MR-DINOSAUR$^*$\textit{(Ours)}                     & RAFT   & 50.8 & 74.7 & 74.9 \\
  \bottomrule
\end{tabularx}

\end{table}
\begin{figure*}[t]    
    \newcommand{\kittiCoef}{0.30}          
\newcommand{\triPdRatio}{0.6038647343} 
\newlength{\kittiWidth}
\setlength{\kittiWidth}{\kittiCoef\textwidth}
\newlength{\triPdWidth}
\setlength{\triPdWidth}
  {\fpeval{\triPdRatio * \kittiCoef}\textwidth} 
\scriptsize
\sffamily
\setlength{\tabcolsep}{1pt}
\renewcommand{\arraystretch}{0.66}
\begin{tabular}{>{\centering\arraybackslash} m{0.02\textwidth} 
                >{\centering\arraybackslash} m{\triPdWidth} 
                >{\centering\arraybackslash} m{\triPdWidth} 
                >{\centering\arraybackslash} m{\kittiWidth}
                >{\centering\arraybackslash} m{\kittiWidth}}

& \multicolumn{2}{c}{\textbf{TRI-PD}} & \multicolumn{2}{c}{\textbf{KITTI}} \\
\cmidrule(l{0.5em}r{0.5em}){2-3} \cmidrule(l{0.5em}r{0.5em}){4-5}

\rotatebox[origin=lB]{90}{\scriptsize{\hspace{-28.8em}\MethodName\hspace{1.4em}DIOD\hspace{2.7em}DINOSAUR\hspace{1.3em}Ground~truth\hspace{2.2em}Image}} &
\includegraphics[width=\linewidth]{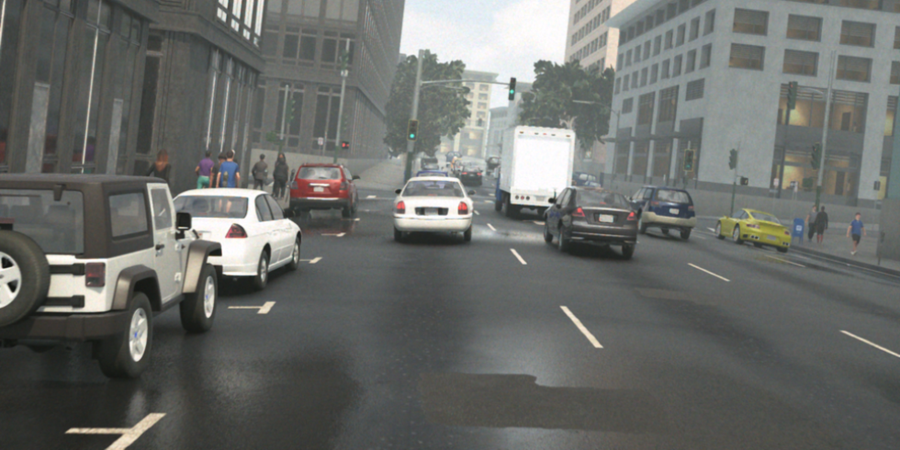}  &
\includegraphics[width=\linewidth]{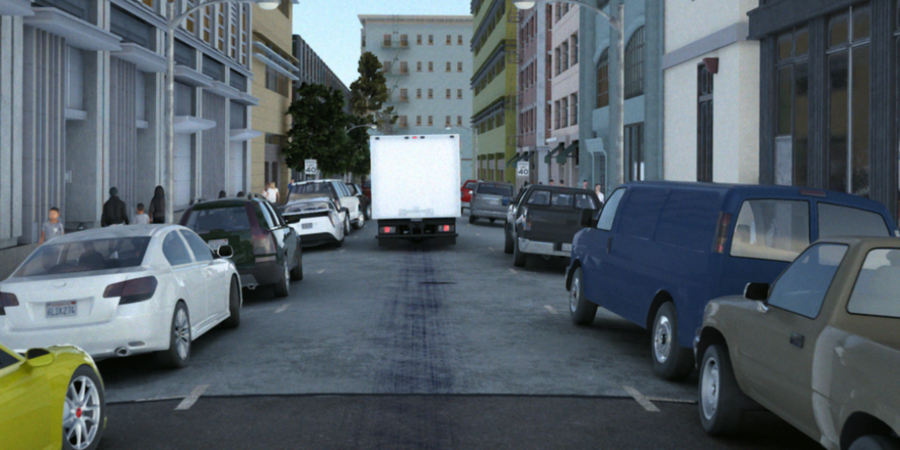}&
\includegraphics[width=\linewidth]{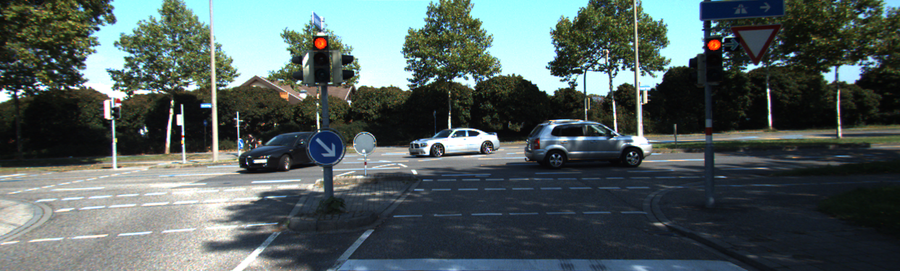}  &
\includegraphics[width=\linewidth]{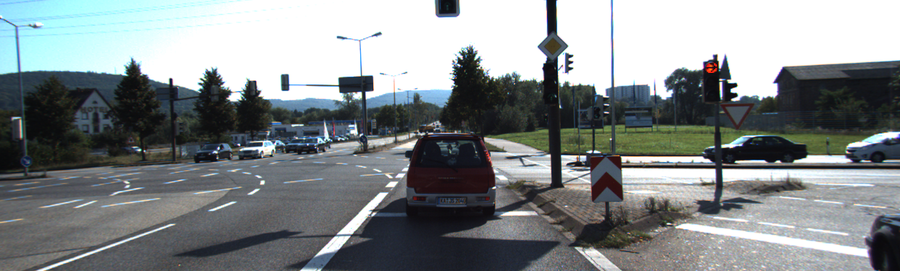}  \\

&
\includegraphics[width=\linewidth]{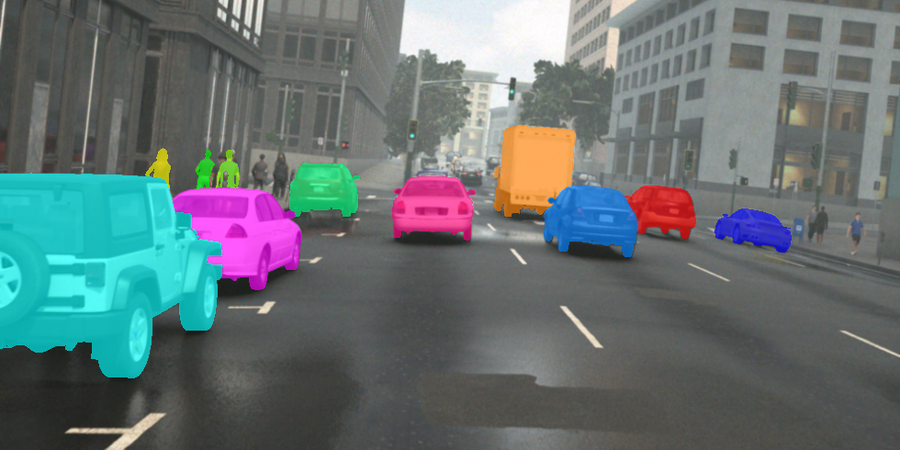}  &
\includegraphics[width=\linewidth]{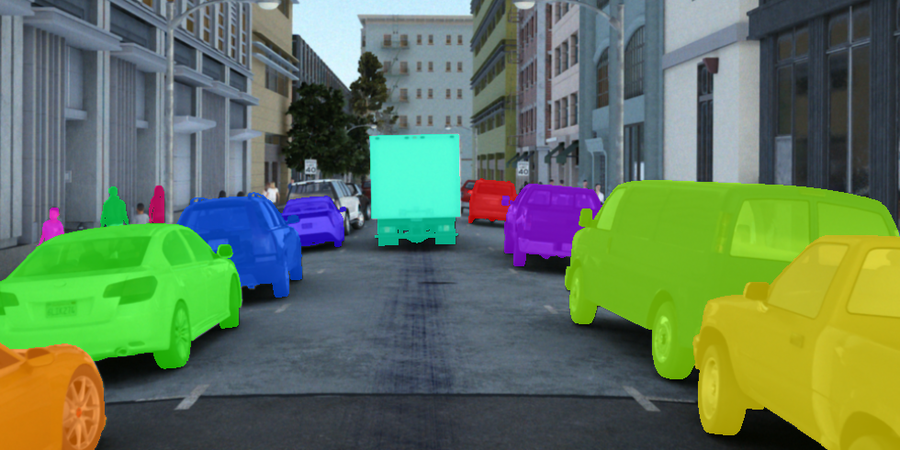} & \includegraphics[width=\linewidth]{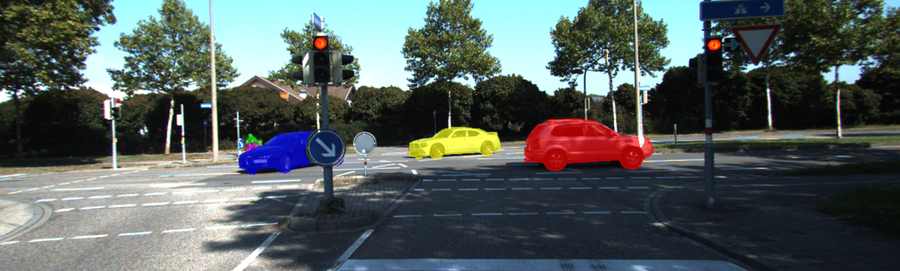}  &
\includegraphics[width=\linewidth]{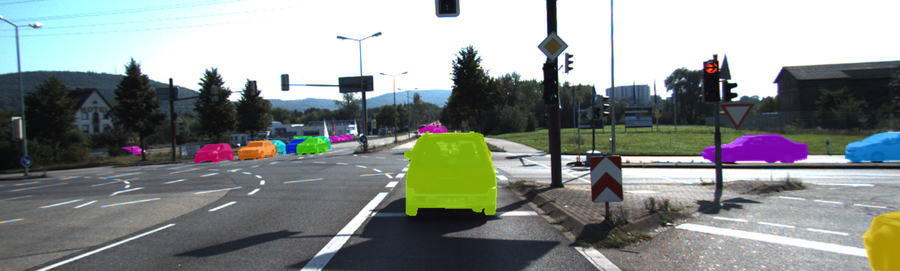}   \\

& 
\includegraphics[width=\linewidth]{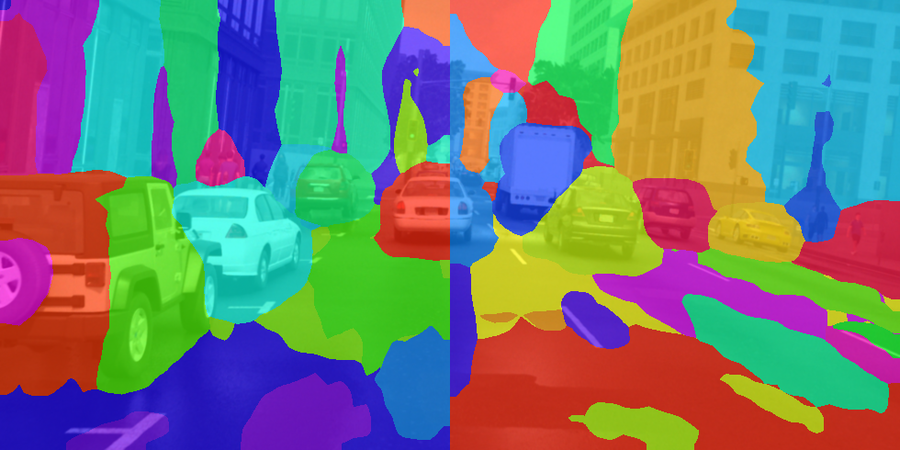}  &
\includegraphics[width=\linewidth]{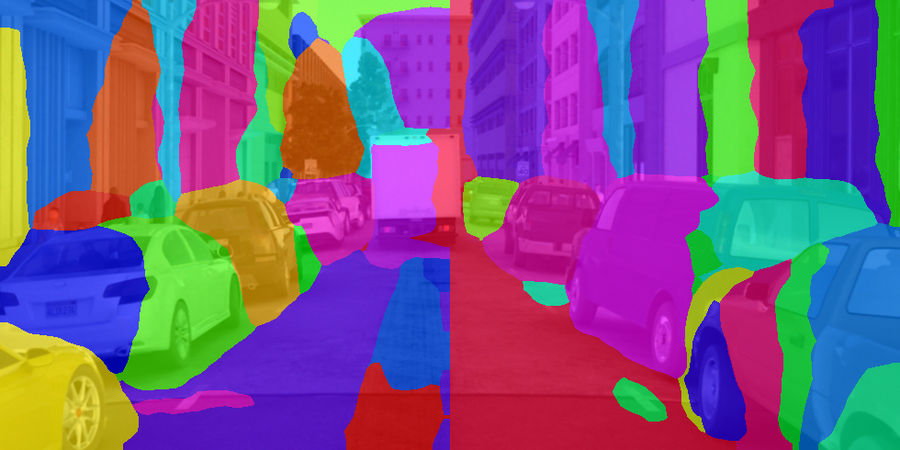} &\includegraphics[width=\linewidth]{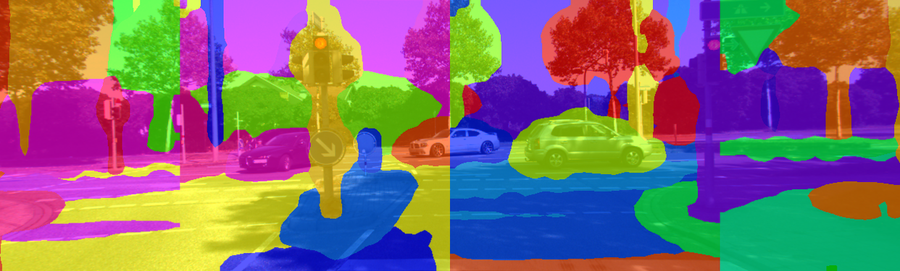}  &
\includegraphics[width=\linewidth]{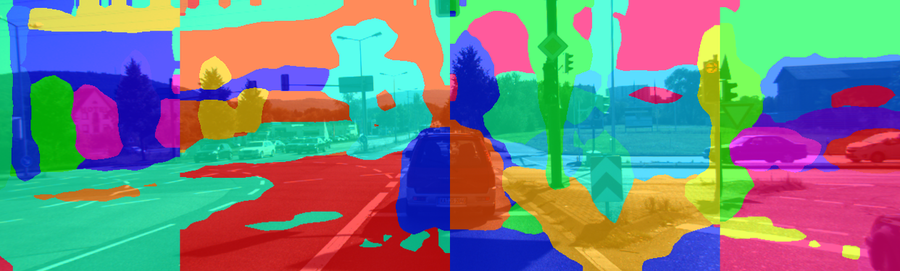}    \\

&
\includegraphics[width=\linewidth]{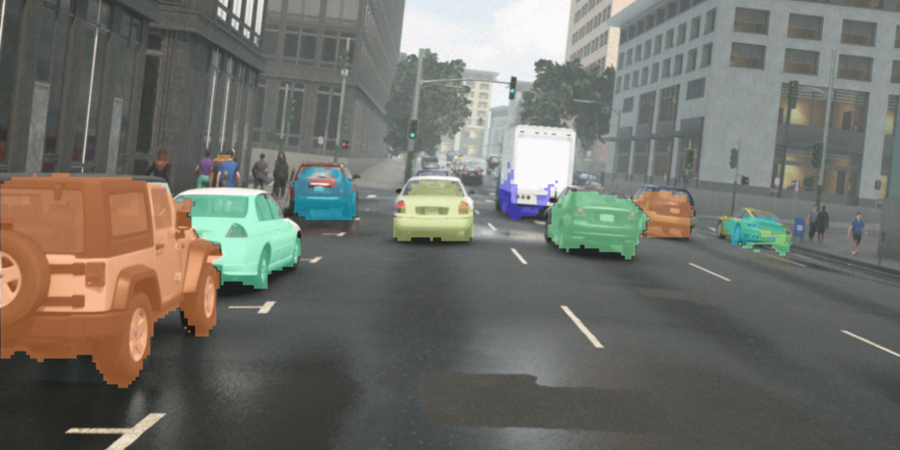}  &
\includegraphics[width=\linewidth]{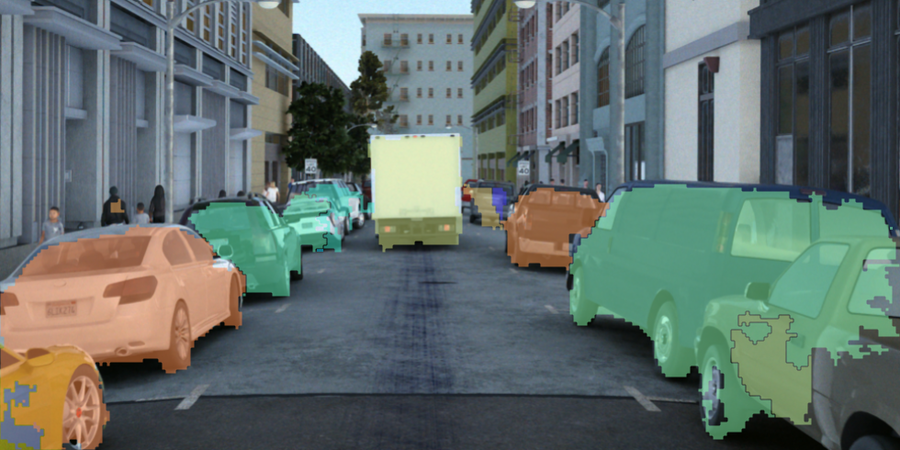} & \includegraphics[width=\linewidth]{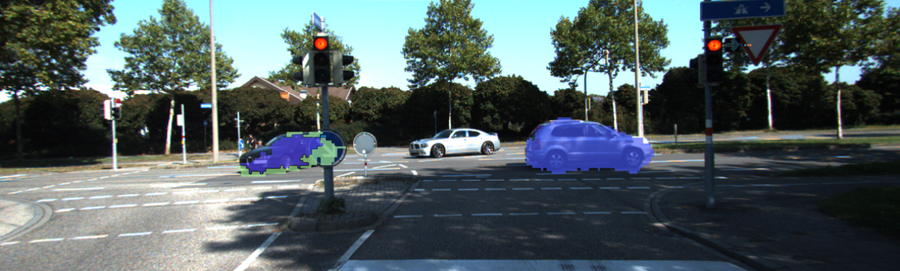}  &
\includegraphics[width=\linewidth]{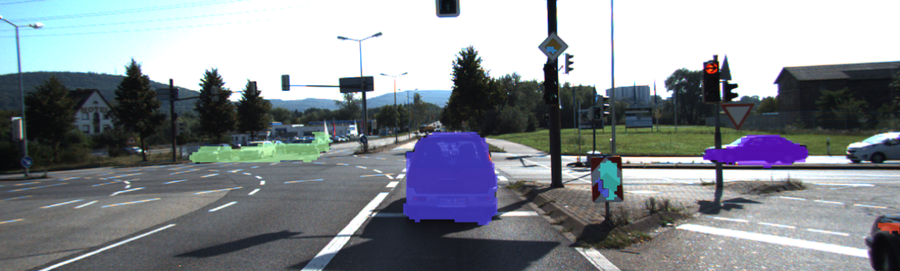}    \\

& 
\includegraphics[width=\linewidth]{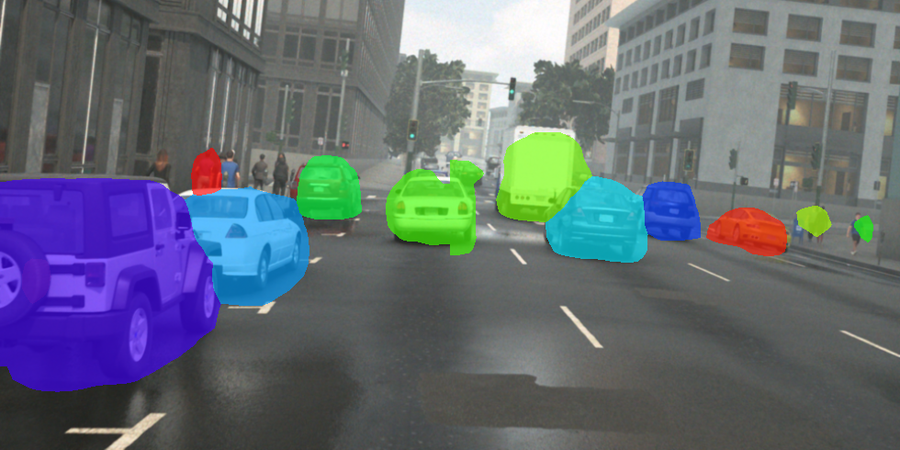}  &
\includegraphics[width=\linewidth]{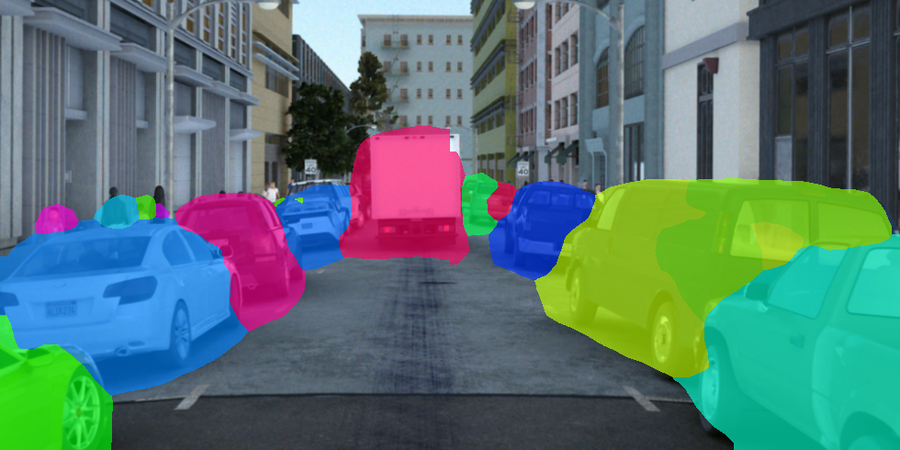} &\includegraphics[width=\linewidth]{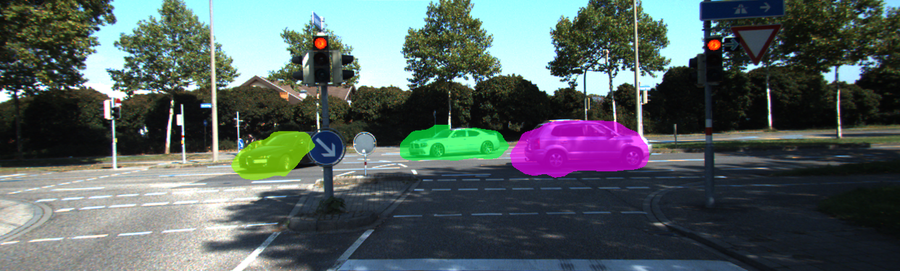}  &
\includegraphics[width=\linewidth]{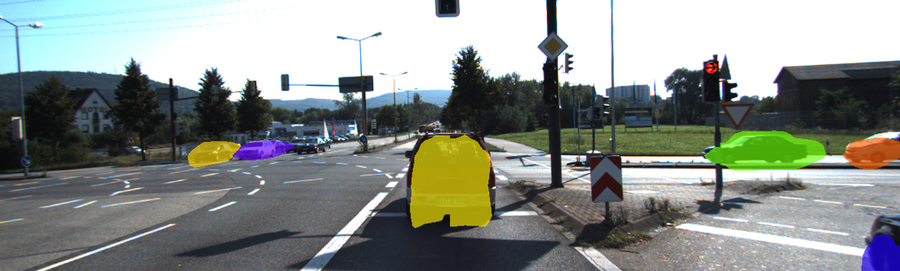}   \\ [-5pt]
\end{tabular}

    \caption{\textbf{Qualitative comparison} of our baseline DINOSAUR~\cite{Seitzer:2023:BGR}, DIOD~\cite{Kara:2024:DSD}, and \MethodName \textit{(Ours)} on the TRI-PD~\cite{Bao:2022:DOT} and  KITTI~\cite{Geiger:2013:VMR} datasets.  We use random colors for different objects. \label{fig:qualitative}}
    \vspace{-0.5em}
\end{figure*}
In \cref{tab:raft_training}, we train our method using pseudo-labels generated with RAFT. This yields comparable fg-ARI values and improvements in all-ARI and F1\textsubscript{50}. 
Notably, using supervised RAFT flow, as done by DIOD, leads to state-of-the-art results on the all-ARI metric as well.

\subsection{Qualitative results} 
We show qualitative results of our method \MethodName and the baselines on both TRI-PD and KITTI datasets in \cref{fig:qualitative}. 
Compared to DIOD~\cite{Kara:2024:DSD}, we observe that \MethodName leads to less noisy predictions, discovering a large number of objects while rarely predicting masks on the image background. \MethodName predicts coarser masks compared to DIOD, which we attribute to the nature of the DINOSAUR slots. We discover more instances compared to DIOD, even for finely resolved distant objects. 

\section{Limitations and Future Work \label{sec:future_work}}
Our pseudo-labels result from the static background assumption in our motion segmentation method. We rely on retrieving quasi-static frames and require training data captured with a static camera---a condition met in most but not all real-world scenarios. Additionally, using unsupervised flow estimation introduces apparent motion artifacts (\eg, a car's shadow in \cref{fig:pseudo_qualitative}) in the pseudo-labels. 
More generally, current motion-based multi-object discovery approaches share a limitation: learning the concept of an object from motion is, by definition, restricted to object categories capable of moving~\cite{Choudhury:2022:GWM, Safadoust:2023:MOD, Kara:2024:DSD}. In contrast, a related line of research~\cite{Wang:2023:TSO, Wang:2023:CLU, Wang:2024:VSS} leverages pseudo-labels from the object-centricity of datasets (\eg, ImageNet~\cite{Russakovsky:2025:ILS}) to learn instances without supervision. Combining these strengths could offer a more comprehensive solution for multi-object discovery.
In general, our method is agnostic to the baseline OCL approach, and could benefit from future, more advanced OCL approaches. In summary, we propose a minimalistic, fully unsupervised method for multi-object discovery that can be a foundation for advancing instance-level understanding.

\section{Conclusion\label{sec:conclusion}}
We presented \MethodName, extending the established OCL approach DINOSAUR for the task of unsupervised multi-object discovery.
We demonstrate that selecting quasi-static video frames enables the generation of high-quality instance pseudo-labels without any supervision, as it simplifies the problem to clustering optical flow of moving objects.
We effectively refined DINOSAUR using the proposed pseudo-labels and extended DINOSAUR for multi-object discovery by introducing a slot deactivation module to distinguish between foreground and background slots.
We propose a unified framework for unsupervised multi-object discovery that requires less data, no supervision, and still achieves strong results across multiple benchmarks compared to previous approaches.

\clearpage
{\small \inparagraph{Acknowledgments.} This project has received funding from the ERC under the European Union’s Horizon 2020 research and innovation programme (grant agreement No.\ 866008). This work has further been co-funded by the LOEWE initiative (Hesse, Germany) within the emergenCITY center [LOEWE/1/12/519/03/05.001(0016)/72] and the Deutsche Forschungsgemeinschaft (German Research Foundation, DFG) under Germany’s Excellence Strategy (EXC 3066/1 “The Adaptive Mind”, Project No. 533717223). This project was also partially supported by the European Research Council (ERC) Advanced Grant SIMULACRON, DFG project CR 250/26-1 ``4D-YouTube'', and GNI Project ``AICC''.  
Christoph Reich is supported by the Konrad Zuse School of Excellence in Learning and Intelligent Systems (ELIZA) through the DAAD programme Konrad Zuse Schools of Excellence in Artificial Intelligence, sponsored by the Federal Ministry of Education and Research. Finally, we acknowledge the support of the European Laboratory for Learning and Intelligent Systems (ELLIS). Special thanks go to Divyam Sheth for his last-minute help with the paper.}
{
    \small
    \bibliographystyle{ieeenat_fullname}
    \bibliography{bibtex/short, references}
}

\clearpage
\clearpage
\setcounter{section}{0}
\renewcommand\thesection{\Alph{section}}
\setcounter{page}{1}
\pagenumbering{roman}
\maketitlesupplementary

We first extend our approach to the synthetic multi-object video dataset MOVI-E~\cite{movi}.
Next, we evaluate the effectiveness of our quasi-static frames retrieval method used for pseudo-label generation.  
Additionally, we provide qualitative insights into our work, including failure cases, pseudo-label visualization, and MR-DINOSAUR results.
We finish with details about the datasets employed, as well as the implementation details, to facilitate reproducibility.

\section{MR-DINOSAUR on MOVI-E}
We experiment on MOVI-E to broaden the range of datasets and methods for comparison. Because none of those methods report F1 or all-ARI on MOVI-E, we restrict our evaluation to fg-ARI here. 
MOVi-E~\cite{movi} introduces a constant, artificial camera motion that violates our static-frame assumption for pseudo-labeling. Despite this disadvantage, our method achieves promising results in the ballpark of methods that explicitly deal with camera motion in the training data as shown in \cref{tab:movi_e_supp}.

\section{Quasi-static Frame Retrieval Analysis}
\label{sec:quasi_static_analyze}

We evaluate the effectiveness of our quasi-static frame retrieval method on the KITTI dataset, which includes detailed annotations of ground truth camera velocity for every video frame. Given that our motion segmentation approach used for pseudo-label generation relies on the static background assumption, the quasi-static frame retrieval plays an important role in achieving high-quality pseudo labels. 
We evaluate the quasi-static frame retrieval by comparing the set of frames our method retrieves from the training data to the ground-truth quasi-static frames.
Ground-truth quasi-static frames are defined as frames with camera velocities below \SI{0.2}{\meter\per\second}. Our method achieves an impressive \SI{99.4}{\%}  accuracy, \SI{99.2}{\%} precision, and \SI{96.6}{\%} recall as shown in \cref{tab:pseudo_label_generalization}, confirming that thresholding the average flow magnitude at the image corners is a simple and effective way to retrieve quasi-static frames.

\begin{table}[t]
\caption{\textbf{Unsupervised multi-object discovery on MOVI-E} using fg-ARI. * denotes using DINOv2. \underline{Underlined methods} use supervision.}
\vspace{-0.5em}
\sisetup{table-number-alignment=center}  
\newcommand{\mytablecolumnwidth}{0.09\linewidth}
\renewcommand{\arraystretch}{1}
\footnotesize
\noindent
\begin{tabularx}{\columnwidth}{>{\hspace{-\tabcolsep}\raggedright\arraybackslash}X 
> {\centering\arraybackslash}p{0.9cm}} 
  \toprule
  \textbf{Method} &\textbf{fg-ARI} \\ 
  \midrule
  \underline{GWM}~\cite{Choudhury:2022:GWM} \scriptsize {{\textcolor{tud0c!80}{BMVC'22}}}  & 42.5 \\ 
  SPOT~\cite{Kakogeorgiou:2024:STW} \scriptsize {\textcolor{tud0c!80}{CVPR'24}} & 59.9  \\
  \underline{PPMP}~\cite{karazija:2022:UMS} \scriptsize {\textcolor{tud0c!80}{NeurIPS'22}}  & 63.1 \\ 
  DINOSAUR~\cite{Seitzer:2023:BGR} \scriptsize {\textcolor{tud0c!80}{ICLR'23}} & 65.1   \\ 
  \underline{MoTok}~\cite{Bao:2023:ODF} \scriptsize {\textcolor{tud0c!80}{CVPR'23}}  & 66.7 \\ 
  \underline{Safadoust et al.}~\cite{Safadoust:2023:MOD} \scriptsize {\textcolor{tud0c!80}{ICCV'23}}  & 78.3 \\ 
  VideoSAUR \cite{videosaur} \scriptsize {\textcolor{tud0c!80}{NeurIPS'23}} & 78.4 \\
  SOLV$^*$  \cite{Aydemir:2023:SOC} \scriptsize {\textcolor{tud0c!80}{NeurIPS'23}}  & 80.8  \\ 
  \underline{DIOD}$^*$ \cite{Kara:2024:DSD} \scriptsize {\textcolor{tud0c!80}{CVPR'24}} & 82.2 \\
  \midrule
  DINOSAUR$^*$~\cite{Seitzer:2023:BGR} \scriptsize {\textcolor{tud0c!80}{ICLR'23}} & 66.2 \\ 
  \rowcolor{tud0c!20}MR-DINOSAUR$^*$ \textit{(Ours)} & 80.1 \\  
  \bottomrule
\end{tabularx}

\label{tab:movi_e_supp}
\end{table}

\begin{table}[t]
    \centering
    \caption{\textbf{Quasi-static frame retrieval analysis} using accuracy, precision, recall (all in \%) on the KITTI dataset. We compare the set of frames retrieved from the training videos by our method to the set of frames with a ground-truth velocity smaller than \SI{0.2}{\meter\per\second}.}
    \label{tab:pseudo_label_generalization}
    \vspace{-0.5em}
    \sisetup{table-number-alignment=center}
\newcommand{\mytablecolumnwidth}{0.2\linewidth}
\footnotesize
\setlength{\tabcolsep}{1pt}
\renewcommand{\arraystretch}{0.99}
\begin{tabularx}{\columnwidth}{
>{\hspace{-\tabcolsep}\raggedright\columncolor{white}[\tabcolsep][\tabcolsep]}
X>{\centering\arraybackslash}
S[table-format=2.1, table-column-width=\mytablecolumnwidth]
S[table-format=2.1,  table-column-width=\mytablecolumnwidth]
@{\hspace{-6pt}}
S[table-format=2.1,  table-column-width=\mytablecolumnwidth]
@{\hspace{-12pt}}}
  \toprule
  {\textbf{Ground-truth Velocity}}&  {\textbf{Accuracy}}  & {\textbf{Precision}} & {\textbf{Recall}}\\
  \midrule
  $<$ \SI{0.2}{\meter\per\second}   & 99.4 & 99.2 & 96.6 \\ 
  \bottomrule
\end{tabularx}

\end{table}

\section{More Qualitative Results}
We provide additional qualitative visualizations of our pseudo-labels and our proposed method \MethodName, as well as failure cases.

\begin{figure*}[t]    
    \newcommand{\kittiCoef}{0.30}          
\newcommand{\triPdRatio}{0.6038647343} 
\setlength{\kittiWidth}{\kittiCoef\textwidth}
\setlength{\triPdWidth}
  {\fpeval{\triPdRatio * \kittiCoef}\textwidth} 
\scriptsize
\sffamily
\setlength{\tabcolsep}{1pt}
\renewcommand{\arraystretch}{0.66}
\begin{tabular}{>{\centering\arraybackslash} m{0.02\textwidth} 
                >{\centering\arraybackslash} m{\triPdWidth} 
                >{\centering\arraybackslash} m{\triPdWidth} 
                >{\centering\arraybackslash} m{\kittiWidth}
                >{\centering\arraybackslash} m{\kittiWidth}}

\rotatebox[origin=lB]{90}{\scriptsize{\hspace{-26.5em}TSAM\hspace{3.3em}Pseudo label\hspace{2.1em}Motion\hspace{4.1em}Image}} & \multicolumn{2}{c}{\textbf{TRI-PD}} & \multicolumn{2}{c}{\textbf{KITTI}} \\
\cmidrule(l{0.5em}r{0.5em}){2-3} \cmidrule(l{0.5em}r{0.5em}){4-5}

& \includegraphics[width=\linewidth]{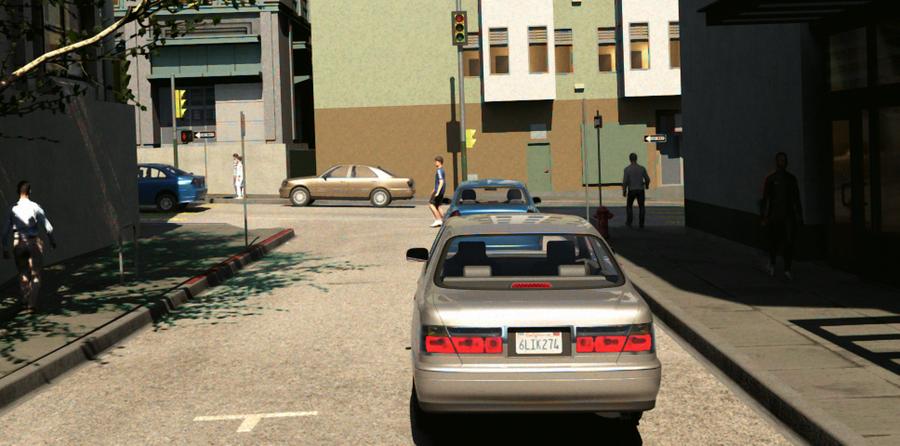}  &
\includegraphics[width=\linewidth]{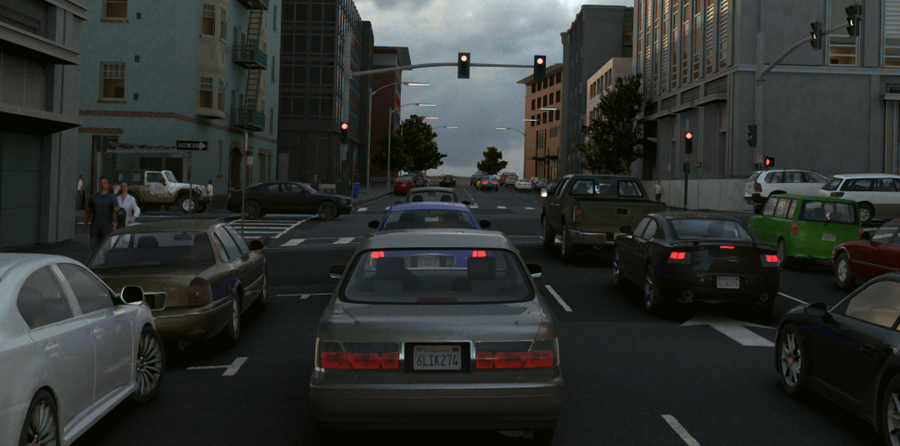} & \includegraphics[width=\linewidth]{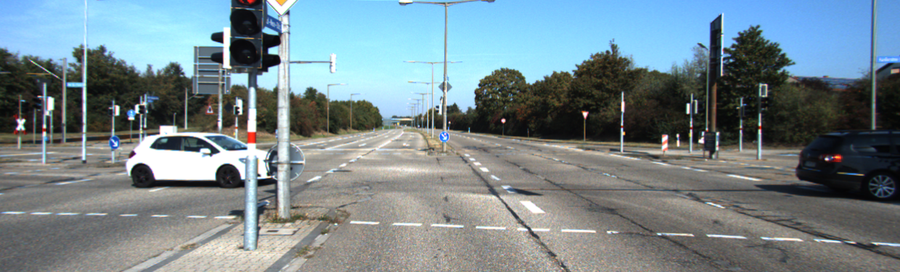}  &
\includegraphics[width=\linewidth]{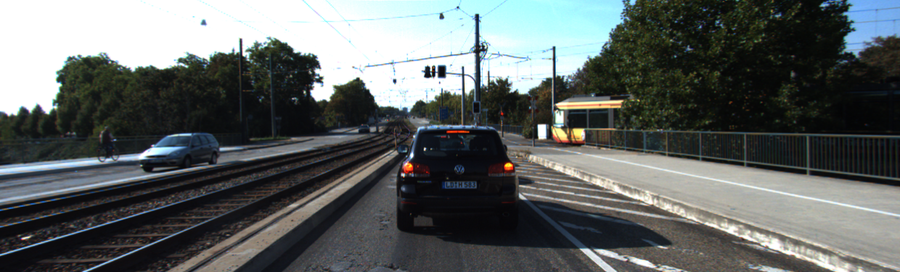}   \\

& \includegraphics[width=\linewidth]{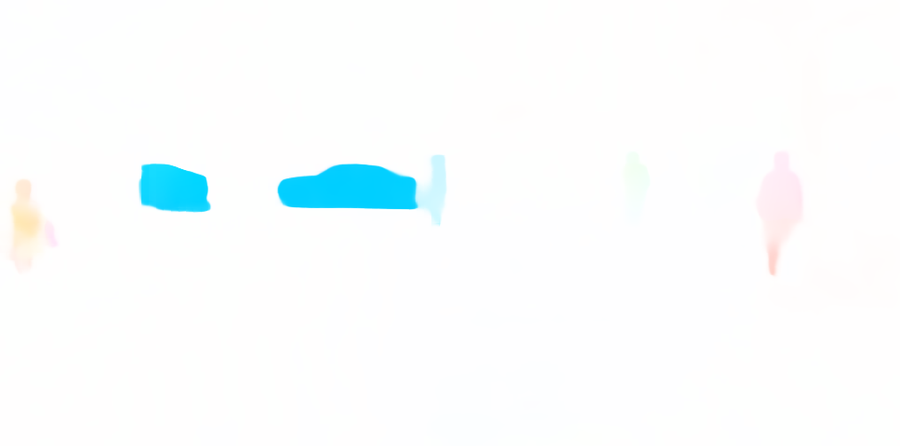}  &
\includegraphics[width=\linewidth]{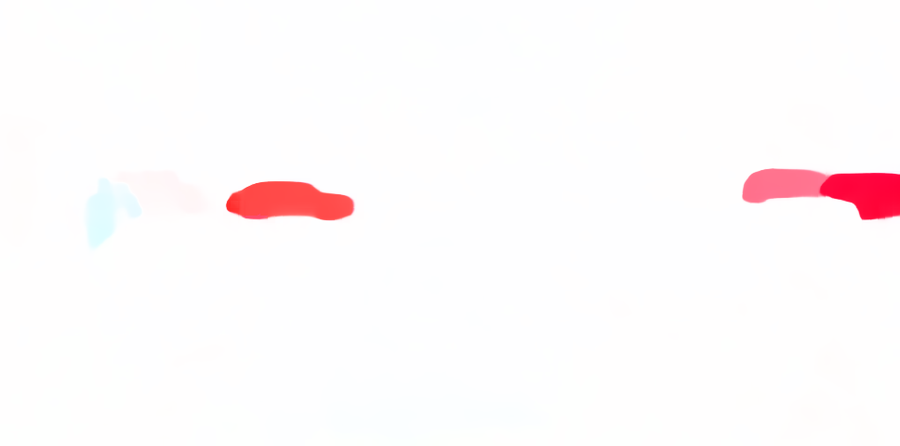}  & \includegraphics[width=\linewidth]{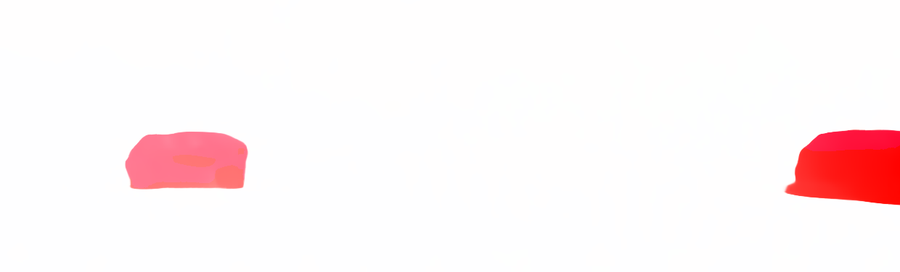}  &
\includegraphics[width=\linewidth]{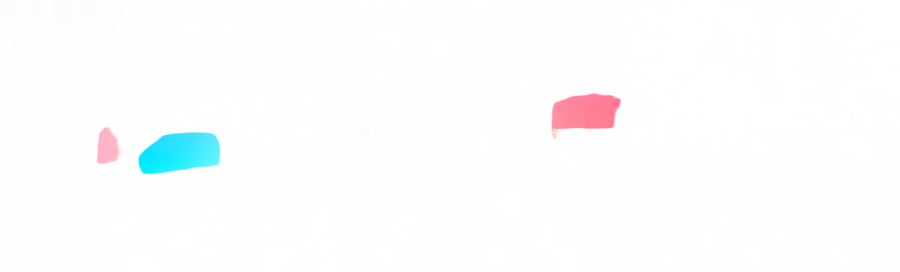} \\

& 
\includegraphics[width=\linewidth]{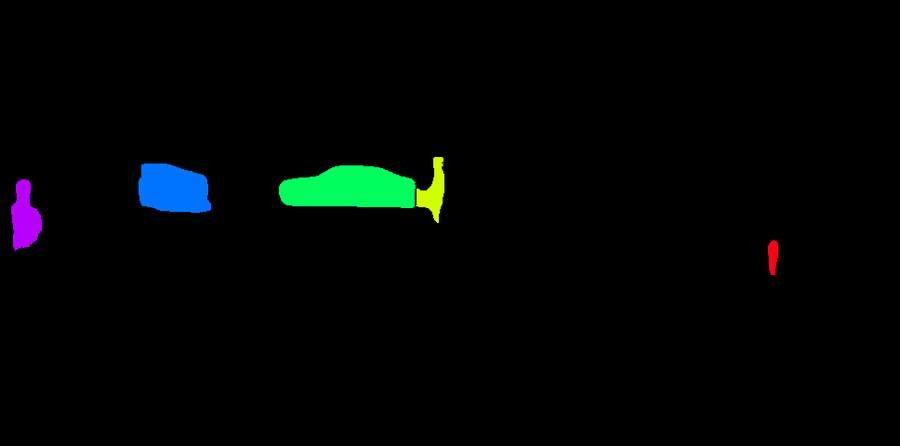}  &
\includegraphics[width=\linewidth]{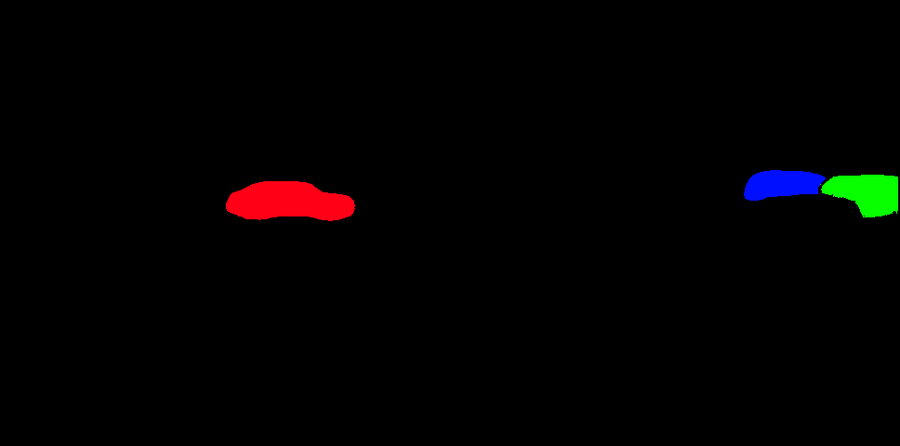} & \includegraphics[width=\linewidth]{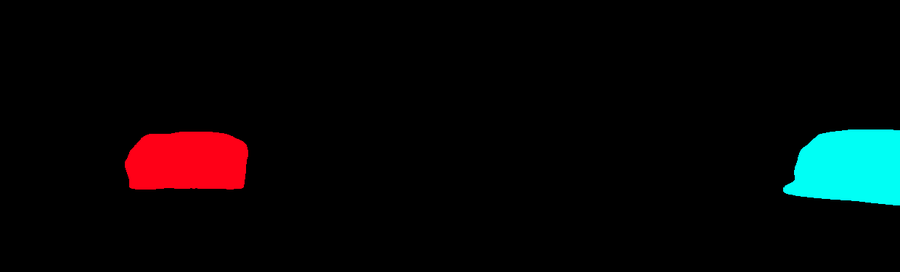}  &
\includegraphics[width=\linewidth]{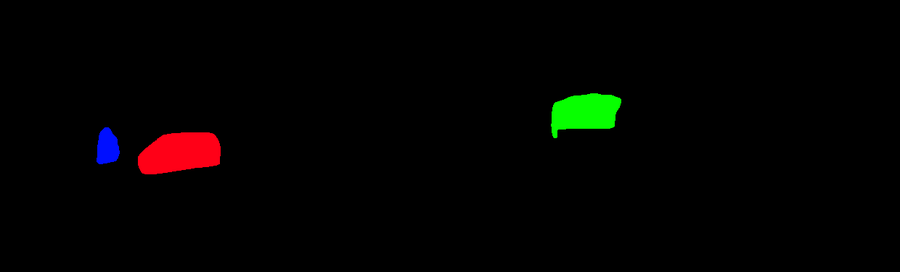}  \\
\midrule

& 
\includegraphics[width=\linewidth]{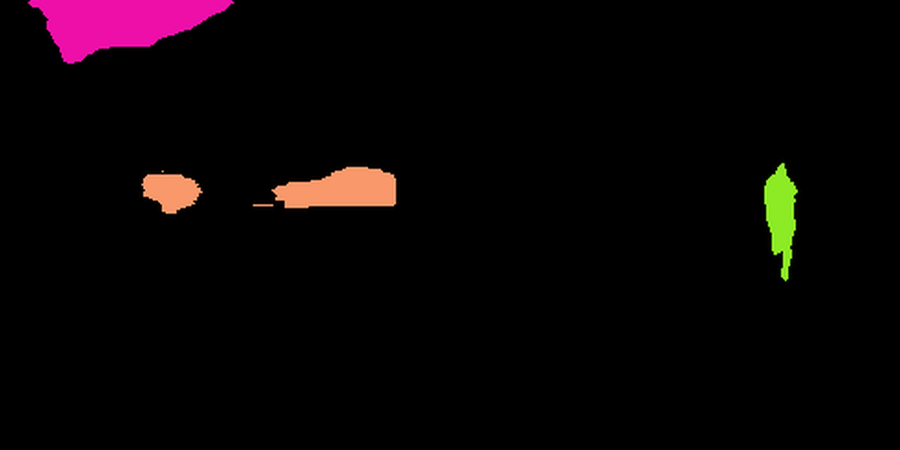}  &
\includegraphics[width=\linewidth]{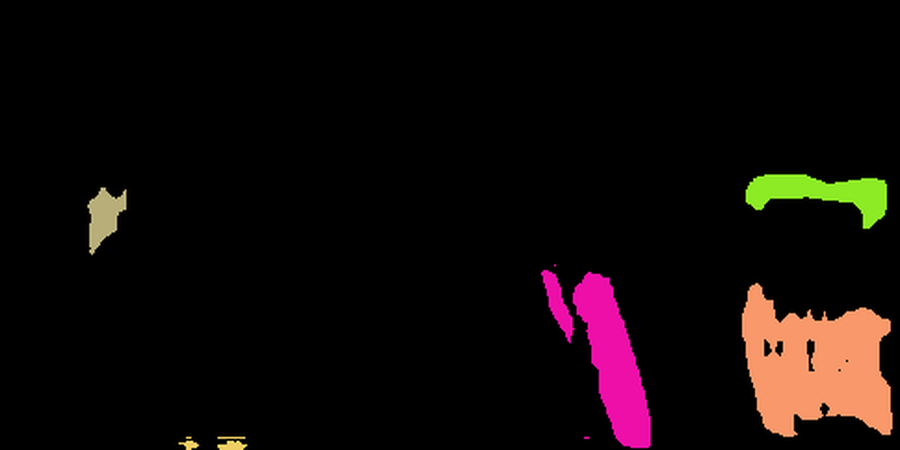}  & \includegraphics[width=\linewidth]{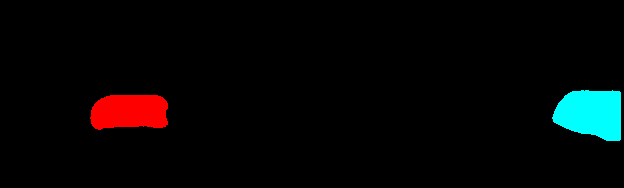}  &
\includegraphics[width=\linewidth]{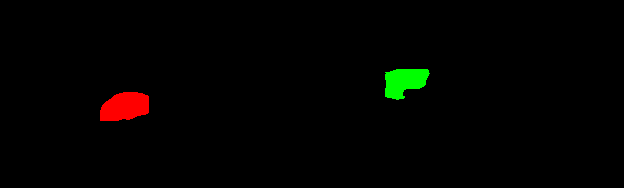}  \\ [-5pt]
\end{tabular}

    \caption{\textbf{Additional visualizations} of our pseudo-labels on the TRI-PD~\cite{Bao:2022:DOT} and  KITTI~\cite{Geiger:2013:VMR} dataset. We further visualize the respective TSAM pseudo labels used by DIOD~\cite{Kara:2024:DSD}. Here we use random colors for different objects. \label{fig:additional_pseudo}}
\end{figure*}

\begin{figure*}[t]    
    \newcommand{\kittiCoef}{0.30}          
\newcommand{\triPdRatio}{0.6038647343} 
\setlength{\kittiWidth}{\kittiCoef\textwidth}
\setlength{\triPdWidth}
  {\fpeval{\triPdRatio * \kittiCoef}\textwidth} 
\scriptsize
\sffamily
\setlength{\tabcolsep}{1pt}
\renewcommand{\arraystretch}{0.66}
\begin{tabular}{>{\centering\arraybackslash} m{0.02\textwidth} 
                >{\centering\arraybackslash} m{\triPdWidth} 
                >{\centering\arraybackslash} m{\triPdWidth} 
                >{\centering\arraybackslash} m{\kittiWidth}
                >{\centering\arraybackslash} m{\kittiWidth}}
                
& \multicolumn{2}{c}{\textbf{TRI-PD}} & \multicolumn{2}{c}{\textbf{KITTI}} \\
\cmidrule(l{0.5em}r{0.5em}){2-3} \cmidrule(l{0.5em}r{0.5em}){4-5}

\rotatebox[origin=lB]{90}{\scriptsize{\hspace{-28.8em}\MethodName\hspace{1.4em}DIOD\hspace{2.7em}DINOSAUR\hspace{1.3em}Ground~truth\hspace{2.2em}Image}} &

\includegraphics[width=\linewidth]{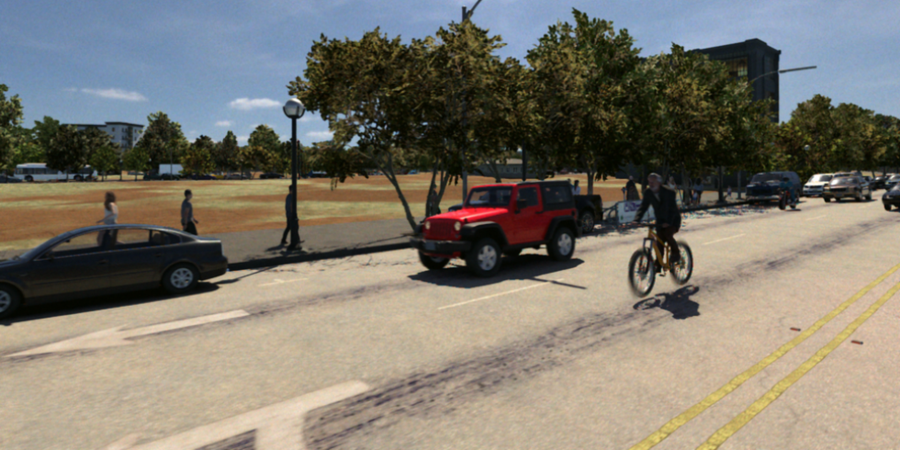}  &
\includegraphics[width=\linewidth]{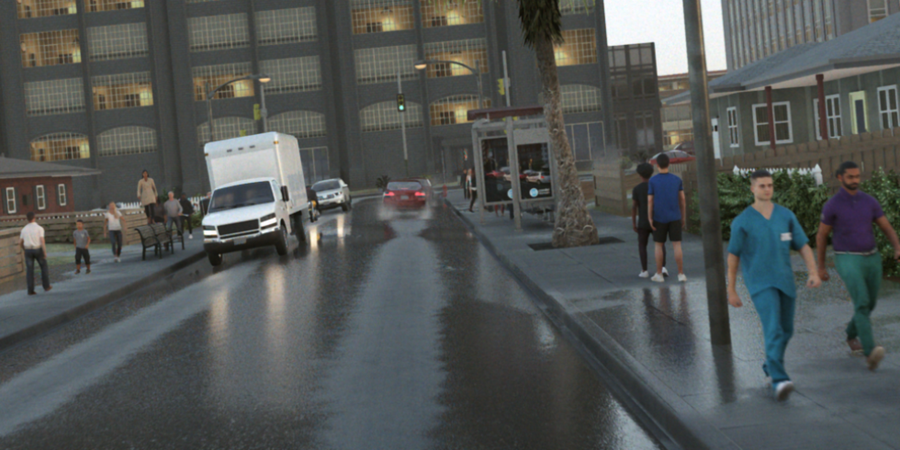}  & \includegraphics[width=\linewidth]{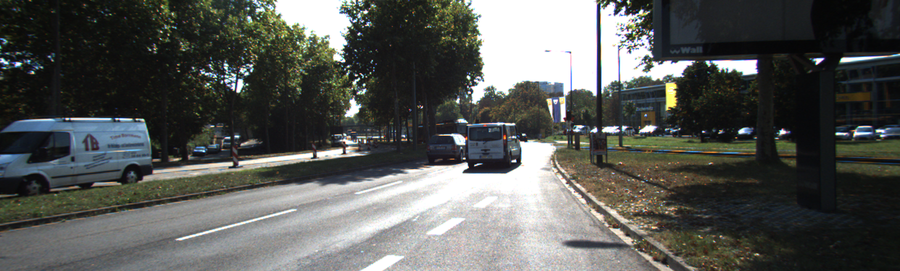}  &
\includegraphics[width=\linewidth]{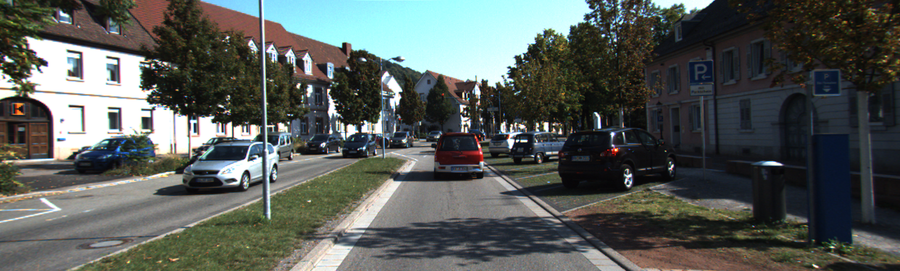}  \\
& 
\includegraphics[width=\linewidth]{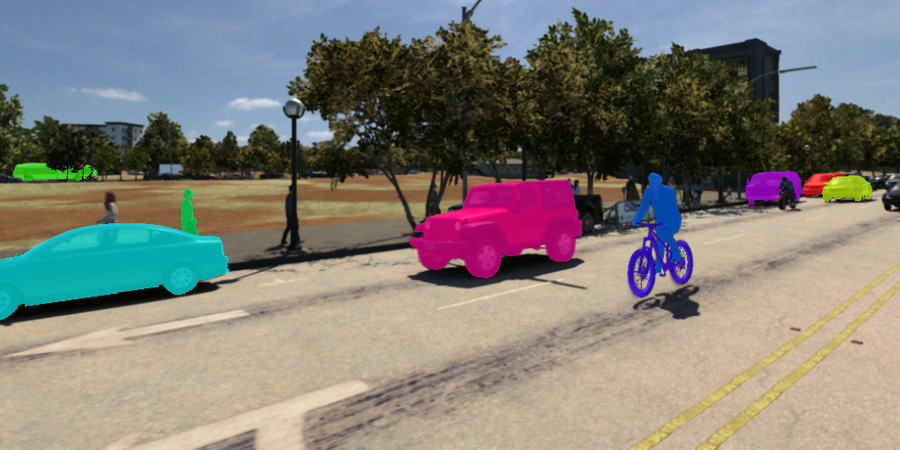}  &
\includegraphics[width=\linewidth]{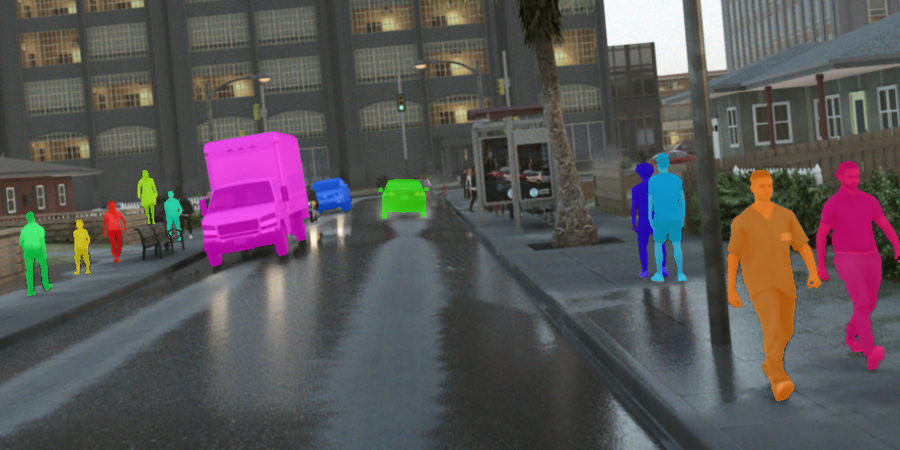} &\includegraphics[width=\linewidth]{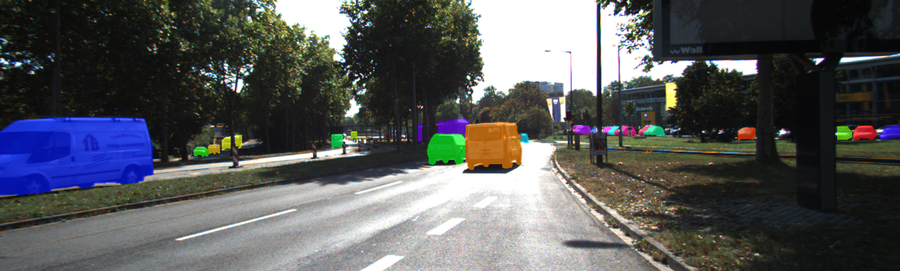}  &
\includegraphics[width=\linewidth]{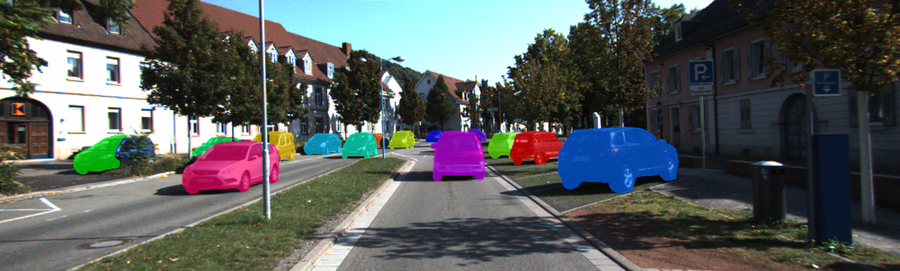}   \\

& 
\includegraphics[width=\linewidth]{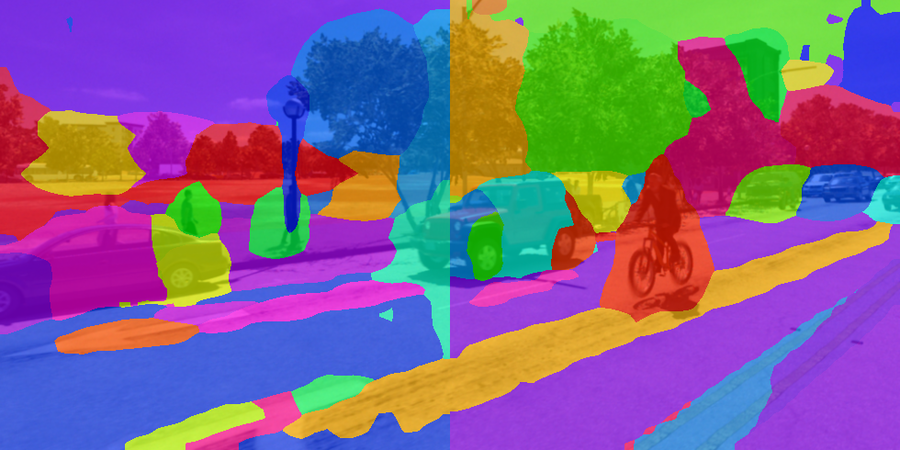}  &
\includegraphics[width=\linewidth]{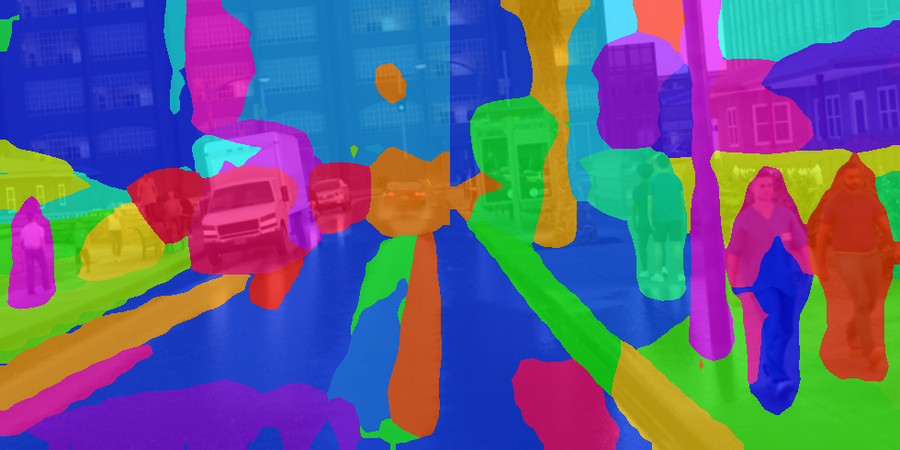} &\includegraphics[width=\linewidth]{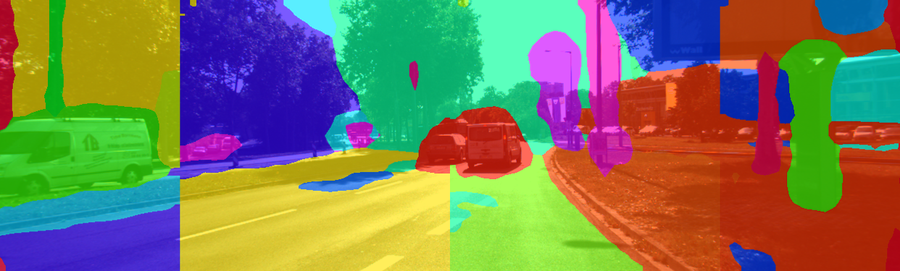}  &
\includegraphics[width=\linewidth]{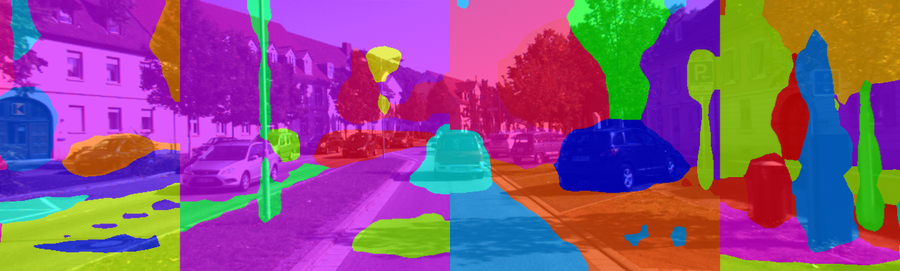}   \\

&
\includegraphics[width=\linewidth]{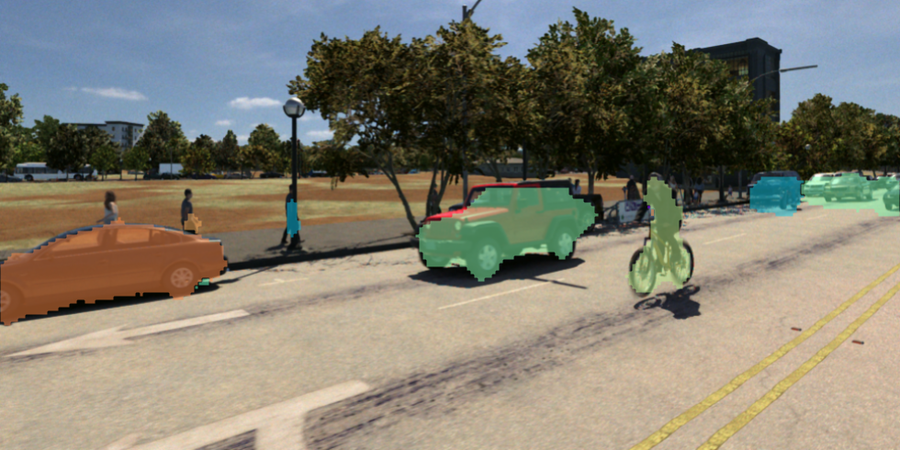}  &
\includegraphics[width=\linewidth]{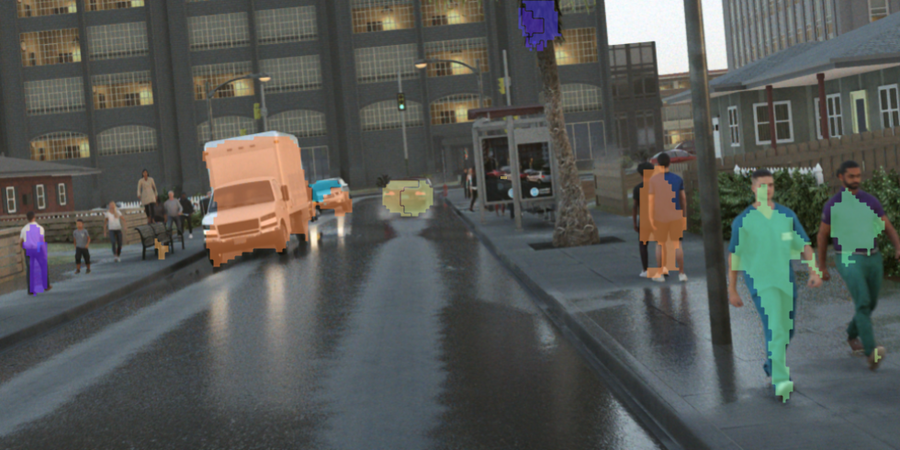} &  \includegraphics[width=\linewidth]{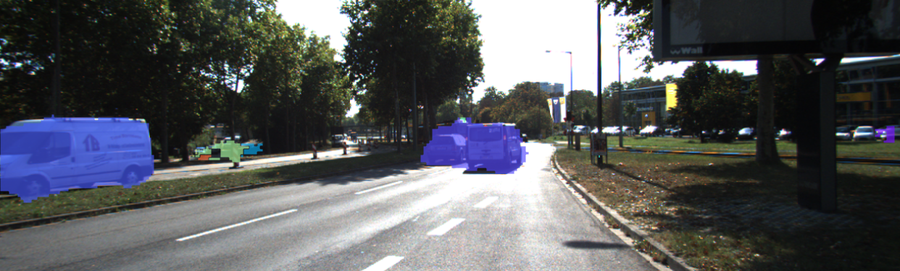}  &
\includegraphics[width=\linewidth]{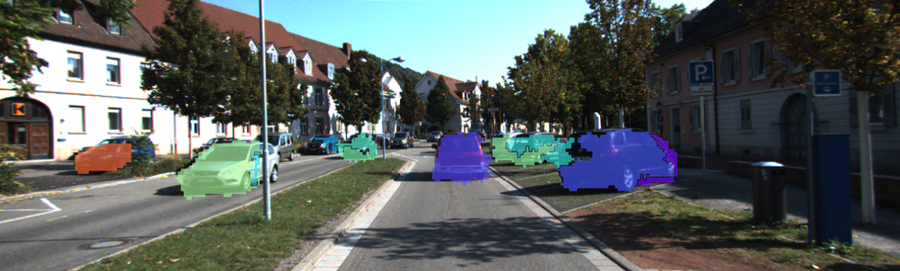}  \\

&
\includegraphics[width=\linewidth]{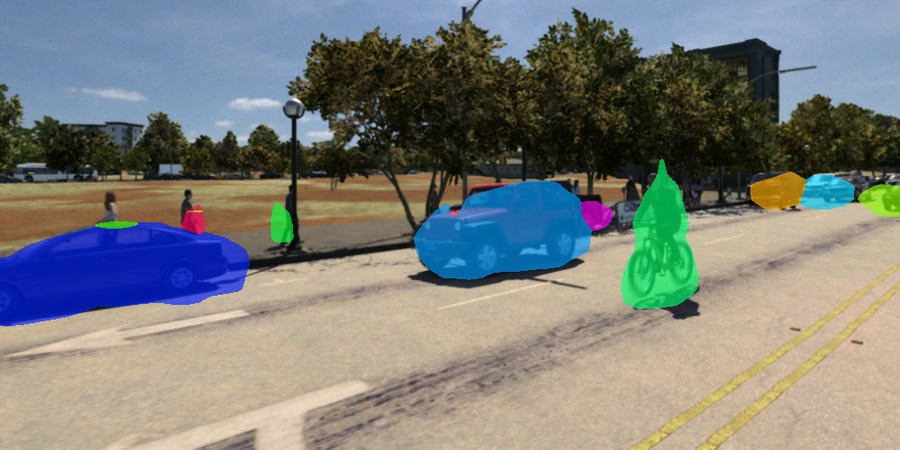}  &
\includegraphics[width=\linewidth]{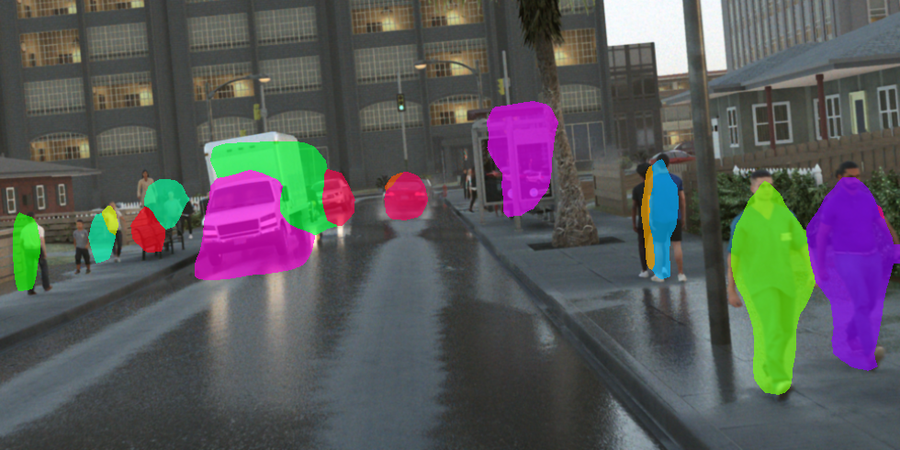} & \includegraphics[width=\linewidth]{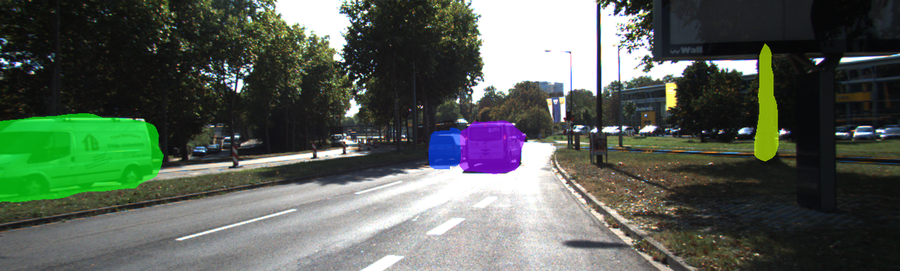}  &
\includegraphics[width=\linewidth]{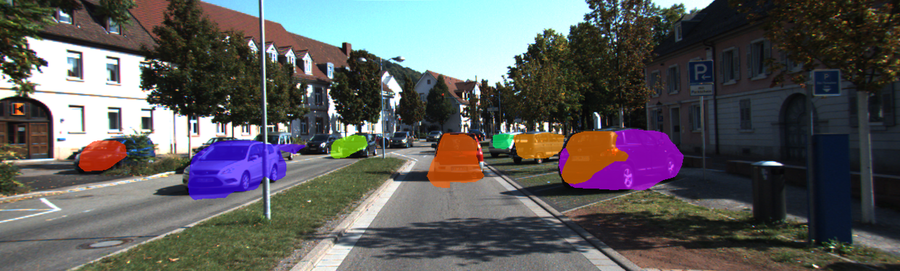}  \\ [-5pt]
\end{tabular}

    \caption{\textbf{Failure cases} of \MethodName \textit{(Ours)} comparing to DIOD~\cite{Kara:2024:DSD} and our baseline DINOSAUR~\cite{Seitzer:2023:BGR} on the TRI-PD~\cite{Bao:2022:DOT} and  KITTI~\cite{Geiger:2013:VMR} dataset. Here we use random colors for different objects. \label{fig:failure_cases}}
\end{figure*}

\subsection{Qualitative pseudo-label examples}
\cref{fig:additional_pseudo} shows additional visualizations of our pseudo-labels comparing to TSAM pseudo labels. Consistent with the analysis in \cref{sec:pseudolabel_analysis}, our pseudo-labels are precise and of high quality for both synthetic TRI-PD and real-world KITTI data. Although some samples exhibit motion artifacts introduced by the unsupervised optical flow from SMURF~\cite{Stone:2021:STM} (\eg, the left KITTI image), we mainly observe accurate object masks. Compared to TSAM pseudo labels, our pseudo labels exhibit fewer artifacts.

\subsection{Failure cases}
We visualize representative failure cases for \MethodName in \cref{fig:failure_cases}. Occasionally, predictions cover non-object structures (\eg, a tree in the left image) and over-segmentation occurs on large objects with intricate textures (\eg, the truck in the right TRI-PD image). Precise segmentation of small, overlapping objects also remains challenging. Notably, similar issues---such as artifacts and missed small objects---are observed with the state-of-the-art DIOD~\cite{Kara:2024:DSD} method.

\begin{figure*}[t]    
    \newcommand{\kittiCoef}{0.30}          
\newcommand{\triPdRatio}{0.6038647343} 

\setlength{\kittiWidth}{\kittiCoef\textwidth}

\setlength{\triPdWidth}
  {\fpeval{\triPdRatio * \kittiCoef}\textwidth}

\scriptsize
\sffamily
\setlength{\tabcolsep}{1pt}
\renewcommand{\arraystretch}{0.66}
\begin{tabular}{>{\centering\arraybackslash} m{0.02\textwidth} 
                >{\centering\arraybackslash} m{\triPdWidth} 
                >{\centering\arraybackslash} m{\triPdWidth} 
                >{\centering\arraybackslash} m{\kittiWidth}
                >{\centering\arraybackslash} m{\kittiWidth}}
                
& \multicolumn{2}{c}{\textbf{TRI-PD}} & \multicolumn{2}{c}{\textbf{KITTI}} \\
\cmidrule(l{0.5em}r{0.5em}){2-3} \cmidrule(l{0.5em}r{0.5em}){4-5}

\rotatebox[origin=lB]{90}{\scriptsize{\hspace{-28.8em}\MethodName\hspace{1.4em}DIOD\hspace{2.7em}DINOSAUR\hspace{1.3em}Ground~truth\hspace{2.2em}Image}} &
\includegraphics[width=\linewidth]{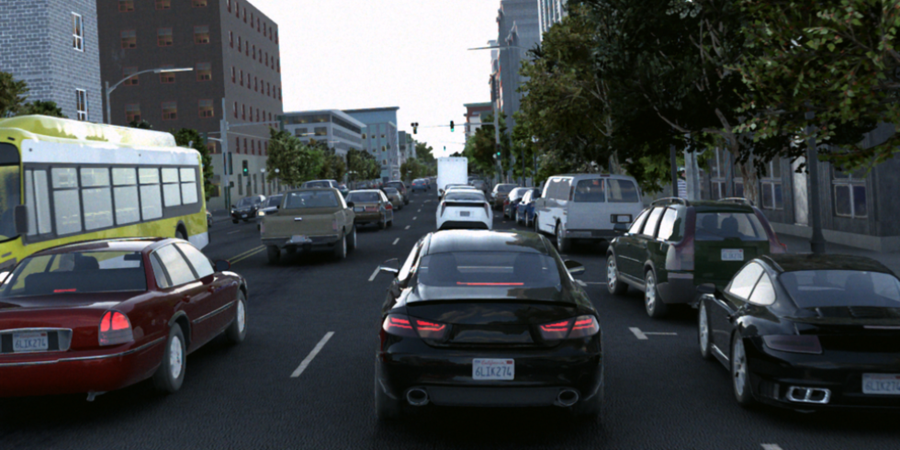}  &
\includegraphics[width=\linewidth]{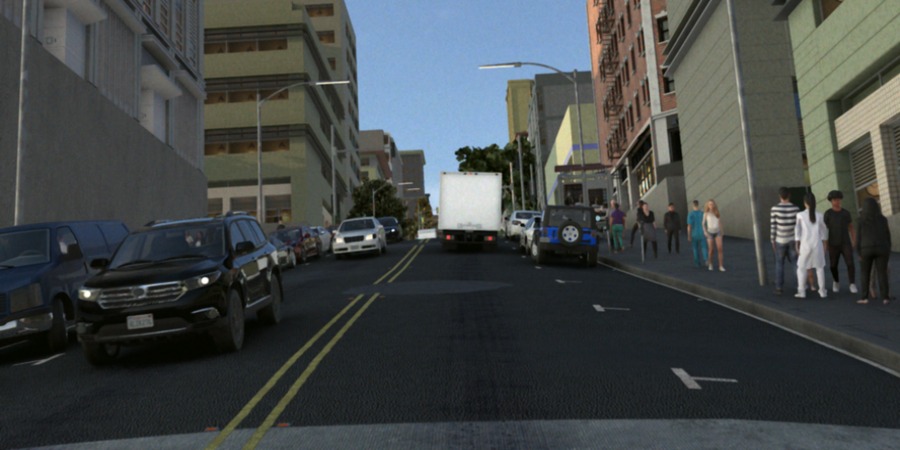} &\includegraphics[width=\linewidth]{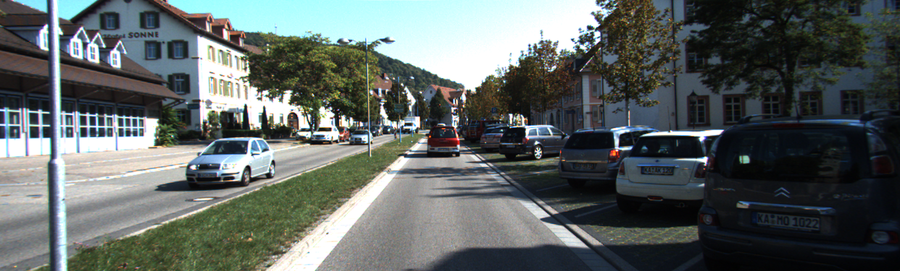}  &
\includegraphics[width=\linewidth]{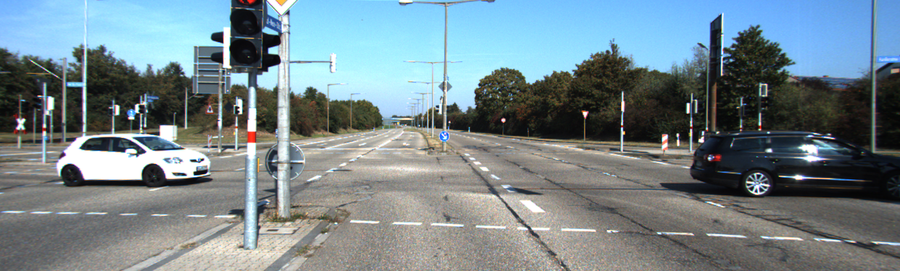}   \\
& 
\includegraphics[width=\linewidth]{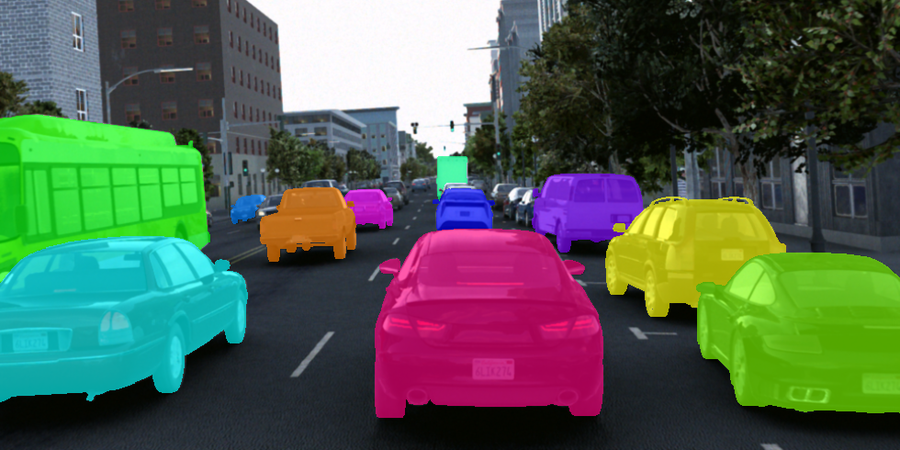}  &
\includegraphics[width=\linewidth]{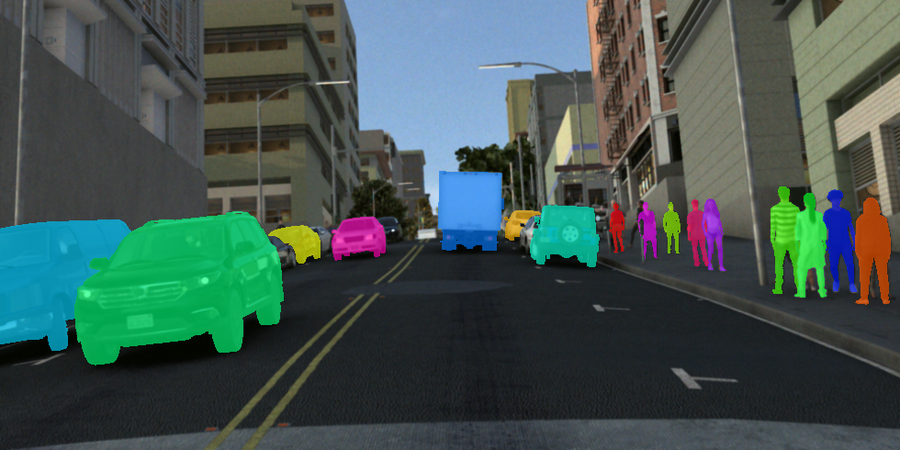} &\includegraphics[width=\linewidth]{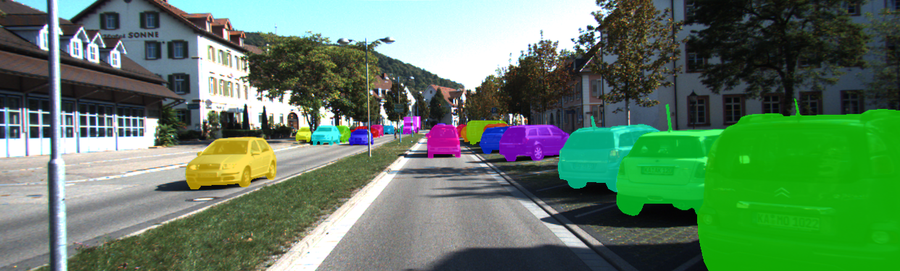}  &
\includegraphics[width=\linewidth]{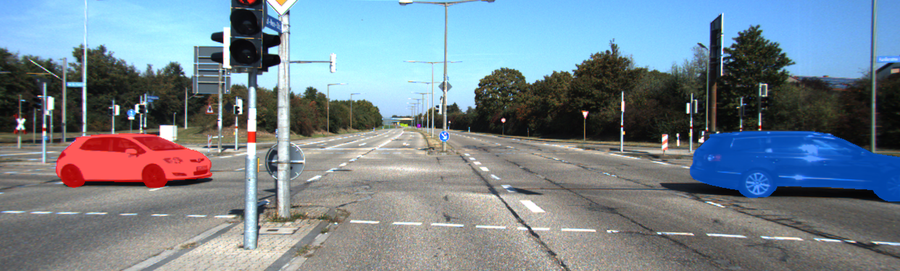}   \\

& 
\includegraphics[width=\linewidth]{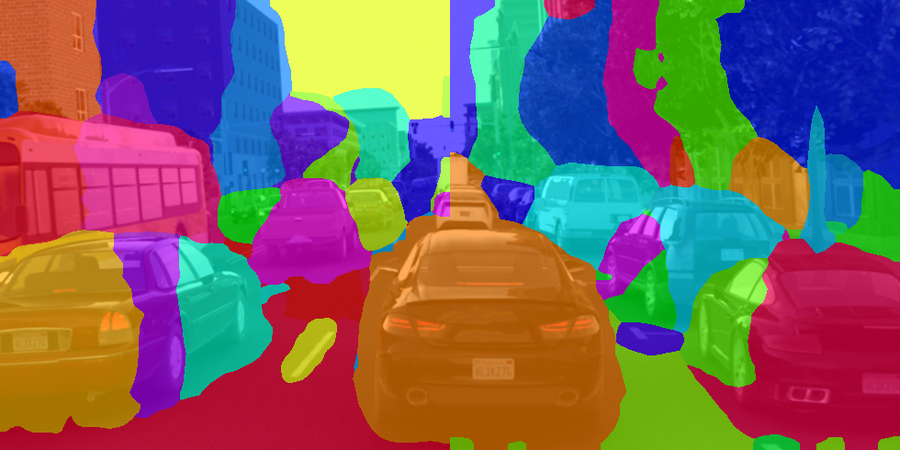}  &
\includegraphics[width=\linewidth]{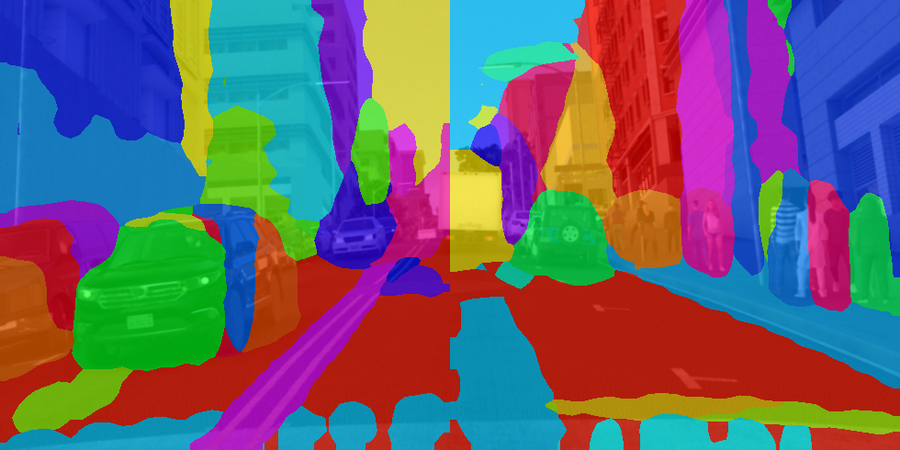} &\includegraphics[width=\linewidth]{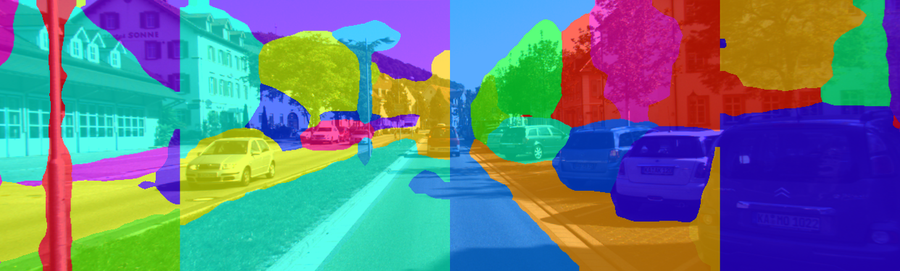}  &
\includegraphics[width=\linewidth]{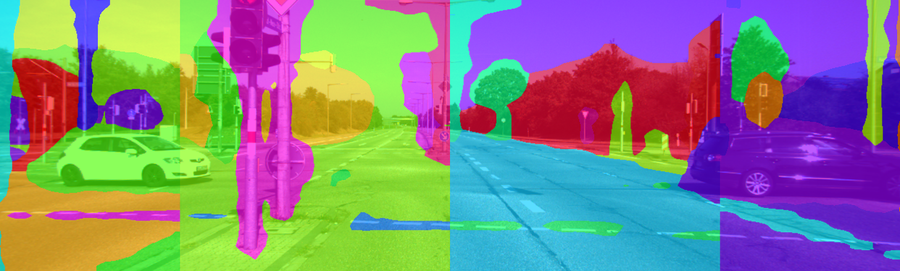}   \\

& 
\includegraphics[width=\linewidth]{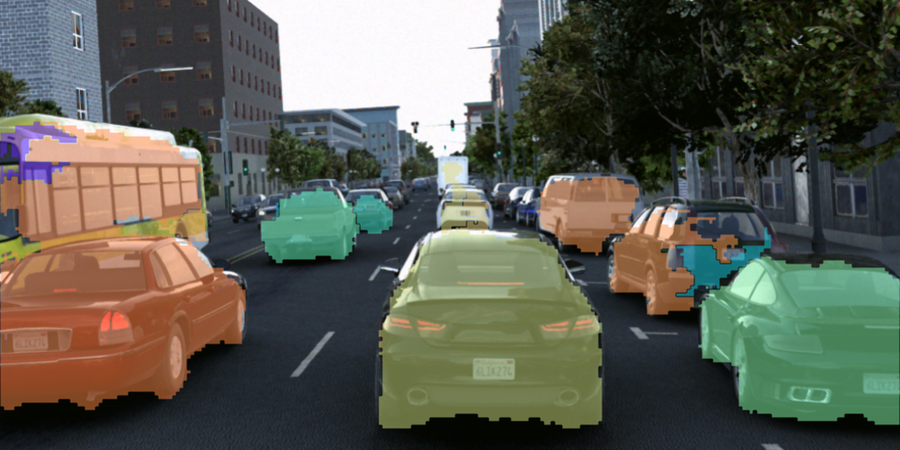}  &
\includegraphics[width=\linewidth]{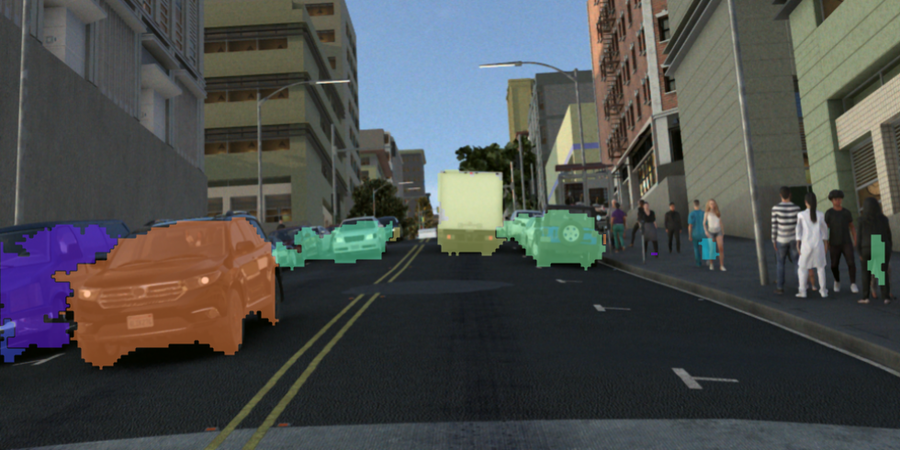} & \includegraphics[width=\linewidth]{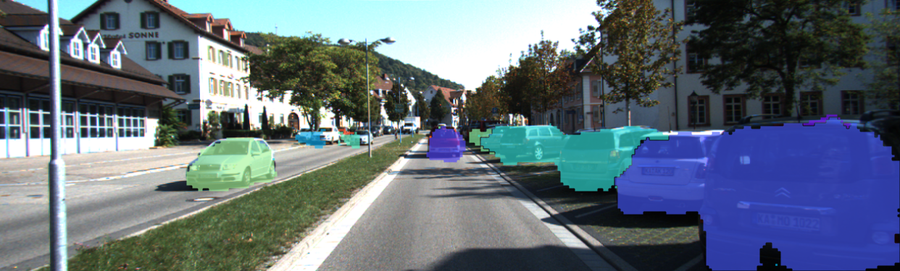}  &
\includegraphics[width=\linewidth]{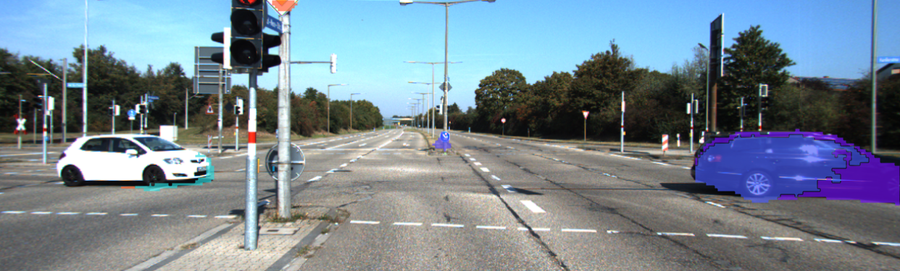}  \\

&
\includegraphics[width=\linewidth]{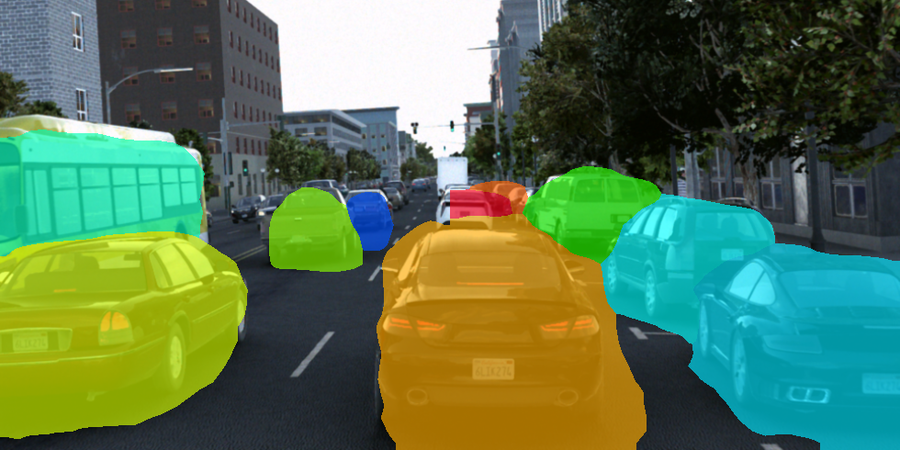}  &
\includegraphics[width=\linewidth]{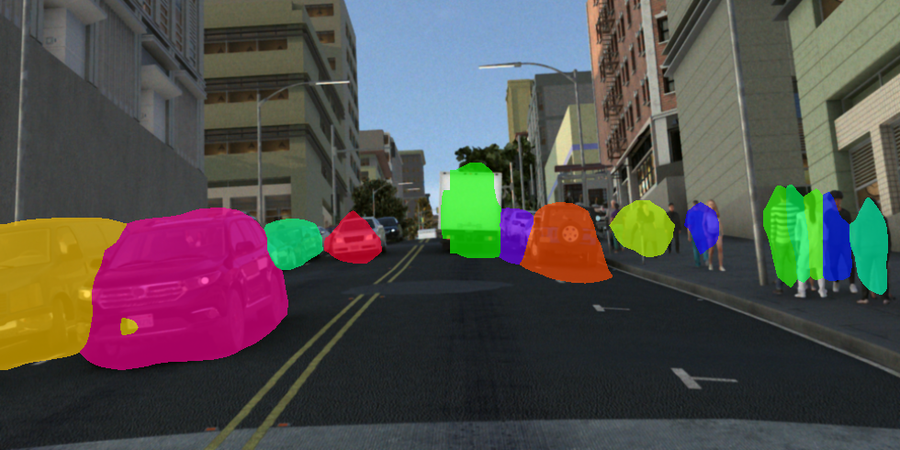} & \includegraphics[width=\linewidth]{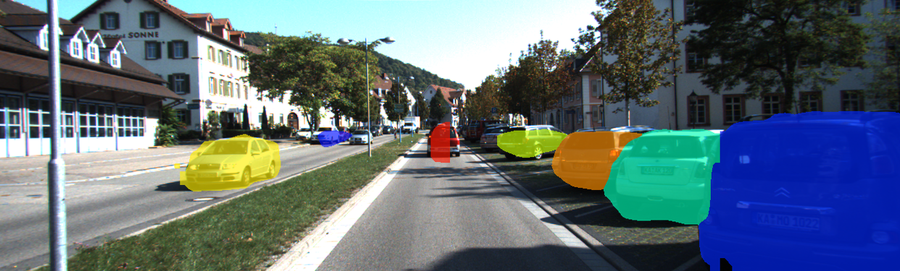}  &
\includegraphics[width=\linewidth]{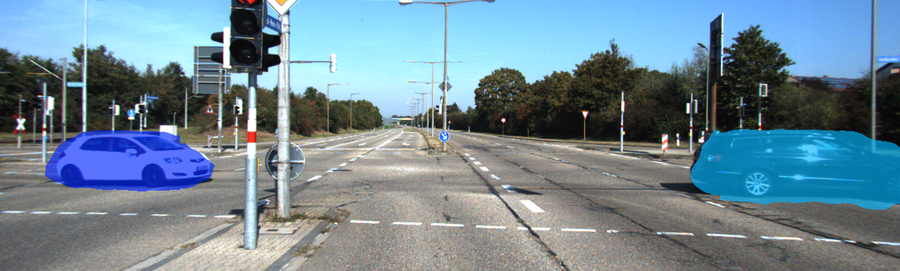}  \\ [-5pt]
\end{tabular}

    \caption{\textbf{Qualitative comparison} of our baseline DINOSAUR~\cite{Seitzer:2023:BGR}, DIOD~\cite{Kara:2024:DSD}, and \MethodName \textit{(Ours)} on the TRI-PD~\cite{Bao:2022:DOT} datasets and  KITTI~\cite{Geiger:2013:VMR}. Here we use random colors for different object instances. \label{fig:addition_qualitative}}
\end{figure*}

\subsection{Qualitative \MethodName examples}
Finally, \cref{fig:addition_qualitative} presents additional qualitative examples comparing our method, \MethodName, to the baseline DINOSAUR~\cite{Seitzer:2023:BGR} and DIOD~\cite{Kara:2024:DSD}. While DINOSAUR establishes a solid foundation, it tends to undersegment and blur the distinction between objects and background. DIOD produces good qualitative results but often yields noisy masks by merging multiple objects into a single mask or missing objects entirely. In contrast, \MethodName effectively differentiates foreground from background, resulting in fewer false positives and demonstrating superior capability in detecting small instances. 

\section{Reproducibility}
To facilitate reproducibility, we elaborate on the technical
and implementation details. Note that our code is available
at \url{https://github.com/visinf/mrdinosaur}.
\subsection{Datasets}

\paragraph{TRI-PD}~\cite{Bao:2022:DOT} is a synthetic urban driving-scene dataset extracted from Parallel Domain~\cite{paralleldomain}. It includes detailed annotations, including camera pose, calibration, depth, instance segmentation, semantic segmentation, 2D/3D bounding box, depth, forward/backward 2D motion vectors, and forward/backward 3D motion vectors. The training set consists of \num{200} photorealistic scenes captured by six cameras, each with \num{200} frames; the validation set comprises \num{17} scenes recorded by three cameras, totaling \num{10200} frames.
In this paper, we only use the three front-camera frames that align with the validation set for training. Following previous work \cite{Bao:2022:DOT, Bao:2023:ODF, Kara:2024:TBA, Kara:2024:DSD}, we discard scenarios with low visibility (\eg, foggy and dark scenes), resulting in \num{157} scenes and a total of \num{94200} frames, for training DINOSAUR. From these, we extract \num{13280} quasi-static frames for training MR-DINOSAUR. 
The resolution of all frames is \num{1216}\,$\times$\,\num{1936}. Following previous work \cite{Bao:2022:DOT, Bao:2023:ODF, Kara:2024:TBA, Kara:2024:DSD}, we resize and crop the images to a resolution of \num{980}\,$\times$\,\num{490} for pseudo-labeling and training. Training is performed on two non-overlapping square crops of size \num{490}\,$\times$\,\num{490}.
\paragraph{KITTI}~\cite{{Geiger:2013:VMR}} is a widely used autonomous driving dataset. It includes various sensor data collected from various environments, \eg, urban, rural, and highway scenes, offering extensive annotations and a diverse range of scenarios. 
We train DINOSAUR using all images provided in the raw data, resulting in  \num{95778} frames from 151 videos. We retrieve \num{12526} quasi-static frames for training MR-DINOSAUR. For evaluation, we utilize the instance segmentation subset, which consists of \num{200} frames with a resolution of \num{375}\,$\times$\,\num{1242}. Each frame is an individual image, rather than part of a consecutive sequence. Also, following previous work~\cite{Bao:2022:DOT, Bao:2023:ODF, Kara:2024:TBA, Kara:2024:DSD}, we resize the images to a resolution of \num{378}\,$\times$\,\num{1260} for pseudo-labeling and training. Training is performed on four non-overlapping square crops of size \num{378}\,$\times$\num{378}.
\paragraph{MOVI}~\cite{movi} is a synthetic video dataset comprising six sub-datasets (MOVI-A to MOVI-F) of increasing complexity. Each sub-dataset consists of generated scenes, with each scene representing a two-second rigid-body simulation of falling objects. The sub-datasets vary in object count and type, background, camera trajectory, and whether all objects are in motion or some remain stationary.
We experiment on the MOVi-E dataset used by several previous works on multi-object discovery \cite{Bao:2023:ODF, Safadoust:2023:MOD, videosaur, Aydemir:2023:SOC, Kara:2024:DSD}. MOVi-E introduces simple camera movement, where the camera moves along a straight line at a random but constant velocity. Each video consists of \num{24} frames, with the training set containing \num{9749} videos (a total of \num{233976} frames) and the validation set containing \num{250} videos (a total of \num{6000} frames). We randomly selected \num{9} frames from each video for training, resulting in a total of \num{87741} images used for training DINOSAUR. We retrieve \num{84831} quasi-static frames for training MR-DINOSAUR. Images originally at \num{256}\,$\times$\,\num{256} are resized to \num{266}\,$\times$\,\num{266} for training to account for the patch size of DINOv2. \\

\begin{table*}[t]
\scriptsize
\setlength{\tabcolsep}{2pt} 
\centering
\caption{DINOSAUR and MR-DINOSAUR hyperparameters used for the results on the TRI-PD, KITTI, and MOVI-E datasets.\label{tab:dinosaur_details}}
\vspace{-0.5em}  
\newcolumntype{R}[1]{>{\raggedleft\arraybackslash}p{#1}}
\begin{tabularx}{\textwidth}{XX R{2.5cm} R{2.5cm} R{2.5cm}}
\toprule
\multicolumn{5}{c}{\textbf{DINOSAUR}}\\
\toprule
\textbf{Dataset} &  &  \textbf{TRI-PD} & \textbf{KITTI} & \textbf{MOVi-E} \\
\midrule
Training steps & & 500k & 500k & 500k \\
Batch size & & 16 & 64 & 64 \\
Optimizer & & Adam & Adam & Adam \\
Number of warmup steps & & 10k &10k & 10k\\
Peak learning rate & & 1e-4&4e-4 & 4e-4 \\
Exponential decay half-life &  & 100k & 100k & 100k \\
ViT architecture &  & DINOv2-ViT-B/14 & DINOv2-ViT-B/14 & DINOv2-ViT-B/14 \\

\midrule
Image/Crop size & & 490 & 378 & 266 \\
Cropping strategy & & Random & Random &  Full\\
Augmentations & & Random Horizontal Flip & Random Horizontal Flip &- \\

\midrule
\multirow{3}{*}{Decoder} & Type & MLP & MLP  & MLP \\
 & Layers & 4 & 4 & 4 \\
 & MLP hidden dimension & 2048 & 2048 & 1024 \\
\midrule
\multirow{4}{*}{Slot Attention} & Number of slots & 30 & 15 & 24 \\
  & Total number of slots & 60 & 60 & 24 \\
 & Iterations & 3 & 3 & 3 \\
 & Slot dimension $D_{slots}$ & 32 & 32 & 128 \\
\toprule
\multicolumn{5}{c}{\textbf{MR-DINOSAUR}}\\
\toprule
\multirow{3}{*}{Pseudo label generation} & Quasi-static frame retrieval threshold $\tau_\text{static}$ & 0.5 & 1.7  & 1.7 \\
 & Foreground mask threshold $\tau_\text{fg}$ & 2.5 & 2.5 & 2.5 \\

 & Flow-gradient threshold $\tau_{\nabla}$ & 20 & 20 & 20 \\
\midrule
\multirow{3}{*}{Training stage 1} & Training epochs & 15 & 15  & 15 \\
 & Batch size & 8 & 8 & 8 \\

 & Learning rate & 4e-06 & 4e-06 & 4e-06 \\

\midrule
\multirow{5}{*}{Training stage 2} & Training epochs & 1 & 1  & 1 \\
 & Batch size & 8 & 8 & 8 \\
 & Learning rate & 4e-05 & 4e-05 & 4e-05 \\
 & Regularization term $\alpha$ & 0.2 & 0.2 & 0.2 \\
 & Drop similarity $\tau_{\text{drop}}$ & 0.99 & 0.99 & 0.99 \\

\midrule
\multirow{2}{*}{Slot deactivation module}  & Layers & 4 & 4 & 4 \\
 & MLP hidden dimension & 2048 & 2048 & 2048 \\

\bottomrule

\end{tabularx}

\end{table*}

\subsection{Computational Requirements}

All experiments use a single NVIDIA RTX \num{6000} Ada Generation GPU (\SI{48}{GB} VRAM) in a workstation equipped with an AMD EPYC \num{7343} CPU (\num{32} cores) and \SI{512}{GB} RAM. 

On TRI-PD, a full \num{500000}-step training of the DINOSAUR baseline takes approximately \SI{267}{h}, involving  \SI{97.3}{M} parameters, of which \SI{10.7}{M} are trainable. For MR-DINOSAUR, training stage 1 (\num{15} epochs, batch size \num{8}) takes approximately \SI{11}{h}, utilizing the same \SI{97.3}{M} parameters, of which \SI{634}{K} are trainable. Training stage \num{2} (\num{1} epoch, batch size \num{8}) takes approximately \SI{40}{min} and utilizes \SI{105.8}{M} parameters with \SI{8.5}{M} trainable. Peak memory usage reaches \SI{40.2}{GB}. At inference, we process an image at a resolution of \num{490}\,$\times$\,\num{980} in \SI{740}{ms}.

\subsection{Further Implementation Details \label{sec:implementation_details}}
Finally, we provide an overview of all hyperparameters used for training the baseline DINOSAUR and our method MR-DINOSAUR in \cref{tab:dinosaur_details}.

\end{document}